\documentclass[10pt,twocolumn,letterpaper]{article}

\usepackage{wacv}              %

\usepackage{subcaption}
\usepackage{adjustbox}
\usepackage{multirow}
\usepackage{pifont}
\usepackage{caption}
\usepackage[table]{xcolor}
\usepackage{xfp}
\usepackage{pgfmath}
\usepackage{enumitem}
\usepackage{float}
\usepackage{algorithm}
\usepackage{algpseudocode}
\usepackage[accsupp]{axessibility}  %
\usepackage{colortbl}

\makeatletter
\DeclareRobustCommand\onedot{\futurelet\@let@token\@onedot}
\def\@onedot{\ifx\@let@token.\else.\null\fi\xspace}

\def\etal{et al\onedot}
\makeatother

\newcommand{\hc}[1]{\cellcolor{green!#1!red!25}#1}
\newcommand{\hcb}[1]{\cellcolor{green!#1!red!25}\textbf{#1}}
\newcommand{\hcu}[1]{\cellcolor{green!#1!red!25}\underline{#1}}
\newcommand{\hcr}[1]{\cellcolor{green!#1!red!25}\rev{#1}}
\newcommand{\hcbr}[1]{\cellcolor{green!#1!red!25}\rev{\textbf{#1}}}
\newcommand{\hcur}[1]{\cellcolor{green!#1!red!25}\rev{\underline{#1}}}

\def\ours{TFD-SfM}

\newif\ifrevmode
\revmodefalse   %
\definecolor{revblue}{rgb}{0,0,0.85}

\makeatletter
\@ifundefined{linelabel}
  {\newcommand{\revlabel}[1]{\label{rev:#1}}}
  {\newcommand{\revlabel}[1]{\linelabel{rev:#1}}}
\makeatother

\newcommand{\rev}[2][]{%
  \if\relax\detokenize{#1}\relax\else\revlabel{#1}\fi
  \ifrevmode\textcolor{revblue}{#2}\else#2\fi}

\newenvironment{revblock}
  {\ifrevmode\color{revblue}\fi}
  {}

\definecolor{wacvblue}{rgb}{0.21,0.49,0.74}
\usepackage[pagebackref,breaklinks,colorlinks,allcolors=wacvblue]{hyperref}

\usepackage{orcidlink}

\def\wacvPaperID{833} %
\def\confName{WACV}
\def\confYear{2027}

\title{Track-Free Dense Structure from Motion}

\author{Pak Hong Chui$^{1,2}$ \quad Hector Andrade-Loarca$^{1,2}$ \quad Daniel Cremers$^{1,2}$\\
$^{1}$Technical University of Munich \quad $^{2}$Munich Center for Machine Learning (MCML)\\
{\tt\small \{p.chui, Hector.Andrade, cremers\}@tum.de}}

\begin{document}
\maketitle

\begin{abstract}
Classical Bundle Adjustment (BA) relies on precise metric initialization and known camera intrinsics, and traditional Structure-from-Motion (SfM) pipelines struggle to exploit modern dense matchers because rigid track-building heuristics break down under their inherent noise. We present Track-Free Dense Structure from Motion (\ours{}), which jointly optimizes camera extrinsics, focal lengths, and geometry from a strict cold start. Our central idea is to relax the rigidities of classical SfM into soft, probabilistic decisions under Negative Log-Likelihood (NLL) objectives. Instead of forcing matched pixels onto a single shared 3D point, we model each match as a 3D Gaussian whose variance is optimized jointly with the geometry. Unreliable matches take on large variance and contribute little to the loss. This smooths the optimization landscape, so the solver converges even from identity cameras and uniform depths. A kinematic relative-pose parameterization stabilizes the cold start, and a dual-hypothesis mechanism resolves the mirror-symmetry ambiguity on object-centric scenes. Across DTU, Mip-NeRF~360, ETH3D, and IMC2021, \ours{} is competitive with both classical and learning-based SfM baselines and leads on ETH3D, recovering accurate camera poses, intrinsics, and dense geometry from unordered, uncalibrated image collections. Project~page: \mbox{\url{https://cphjsn.github.io/tfd-sfm/}}
\end{abstract} %
\section{Introduction}
\label{sec:intro}

\begin{figure}[t]
    \centering
    \includegraphics[width=\linewidth]{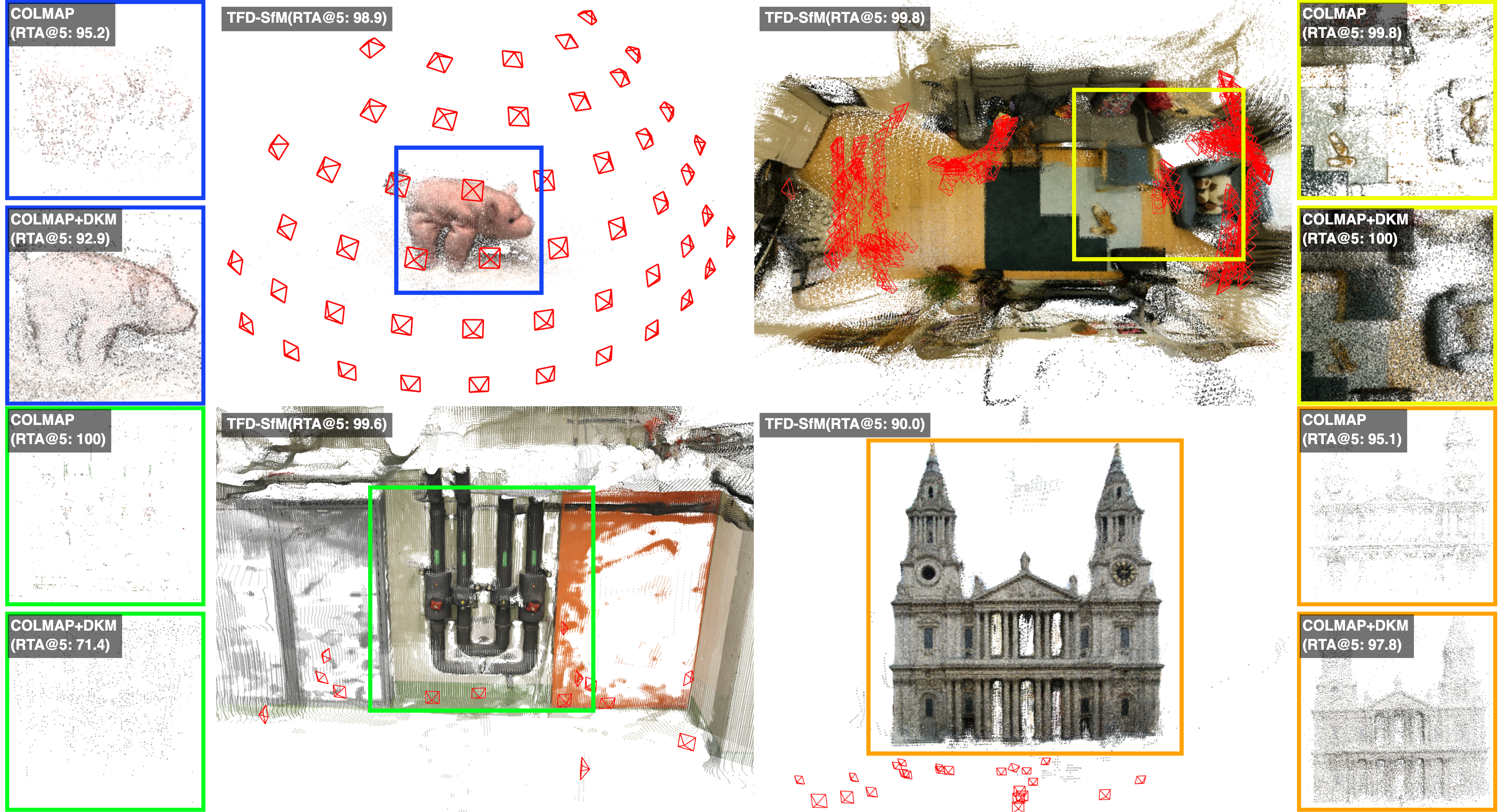}
    \caption{\textbf{Comparison against COLMAP variants.} \ours{} produces robust, dense 3D reconstructions across diverse scenarios: object-centric (blue), wide-baseline indoor (yellow), complex structures (green), and in-the-wild (orange). Gray insets report Relative Translation Accuracy (RTA@5$^\circ$) against standard and dense COLMAP baselines. \ours{} attains competitive metric accuracy with markedly denser structure.}
    \label{fig:teaser}
\end{figure}

Bundle Adjustment (BA) is a core optimization problem in 3D computer vision, providing the standard framework for jointly refining camera poses and scene geometry by minimizing reprojection residuals~\cite{hartley2003multiple, triggs1999bundle}. Beyond its foundational role in Structure-from-Motion (SfM)~\cite{schonberger2016sfm} and Simultaneous Localization and Mapping (SLAM)~\cite{mur2015orbslam, klein2007ptam}, BA has recently become a necessary step for high-fidelity neural rendering techniques, such as Neural Radiance Fields~\cite{mildenhall2021nerf} and 3D Gaussian Splatting~\cite{kerbl20233d}. Despite its success, traditional BA depends on several strong assumptions: the availability of accurate initial pose estimates~\cite{lourakis2009sba}, known camera intrinsics, and reliable correspondences~\cite{triggs1999bundle}. Standard solvers are highly prone to getting stuck in local minima, a problem exacerbated by noise and outliers in the correspondence set~\cite{triggs1999bundle, lourakis2009sba}. These requirements limit the use of BA in unconstrained or unstructured real-world scenarios, where poor initialization or ambiguous geometry often leads to failed or incorrect reconstructions~\cite{torr1999problem, heinly2014correcting}.

\begin{figure*}[t]
    \centering
    \includegraphics[width=\linewidth]{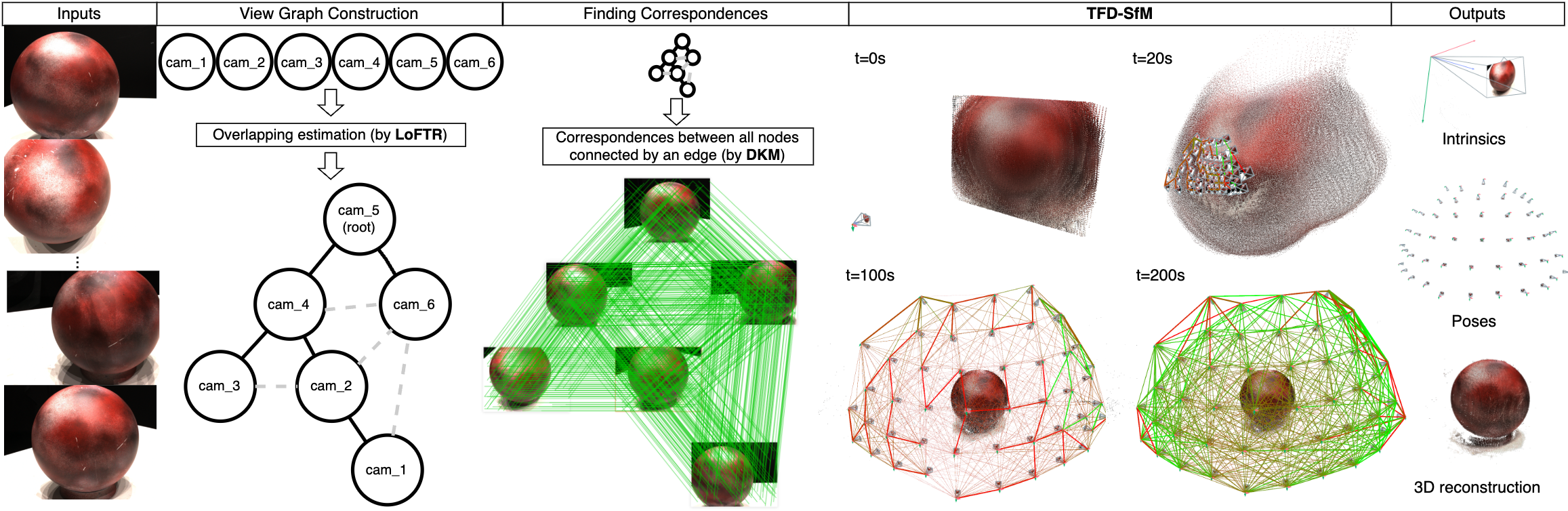}
    \caption{\textbf{System Overview.} From unordered images, three stages: (1) \textbf{View Graph Construction} via LoFTR \cite{sun2021loftr} (solid: spanning-tree edges, dash: auxiliary edges); (2) \textbf{Finding Correspondences} via DKM \cite{edstedt2023dkm}; and (3) \textbf{\ours{} Optimization} from a strict cold start. The temporal sequence ($t=0 \to 200$s) shows the prior-free emergence of the \textbf{outputs}: optimized intrinsics, absolute poses, and a 3D reconstruction.}
    \label{fig:overview}
\end{figure*}

Furthermore, while modern deep networks enable the extraction of highly accurate dense correspondences~\cite{edstedt2023dkm, edstedt2024roma}, effectively leveraging them within classical SfM frameworks remains an unresolved challenge. Traditional pipelines~\cite{schonberger2016sfm, snavely2006photo} rely on the construction of rigid ``tracks'': consistent chains of pixel observations across multiple images assigned to the same 3D point. Naively integrating dense features into systems like COLMAP floods the optimization with noise, causing these fragile track-building heuristics to severely degrade. Learned SfM methods bypass dense matchers and predict their own correspondences, as tracked keypoints in VGGSfM~\cite{wang2024vggsfm} or pairwise pointmaps in DUSt3R~\cite{wang2024dust3r}. They scale poorly to large image sets and remain sensitive to errors in these predictions. Even detector-free SfM~\cite{he2024dsfm}, which replaces the keypoint detector with a dense matcher, still chains its matches into tracks before reconstruction and so remains vulnerable to noise.

To make the BA solver less sensitive to its initialization, recent work has reformulated the cost function to broaden the basin of attraction. One prominent direction involves the use of pseudo-object space errors (\textit{pOSE})~\cite{Hong_2018_CVPR}, which lift 2D observations into a 3D proxy space and apply quadratic regularization to avoid the singularities of the pinhole projection model. Similarly, \textit{expOSE}~\cite{iglesias2023expose} utilizes an exponential depth regularization to prevent geometric degeneracies, while other approaches rely on rotation averaging and translation synchronization to initialize a global coordinate frame~\cite{chatterjee2013efficient, arrigoni2016global}. While these methods reduce trivial solutions by minimizing residuals in the object space, they remain computationally brittle. They frequently rely on manual hyperparameter tuning and fail to reliably resolve the scale-drift~\cite{strasdat2010scale} and mirror-symmetry ambiguities~\cite{ozden2010multibody} inherent in unconstrained, uncalibrated relative pose estimation.

\medskip\noindent\textbf{Our approach.} We introduce Track-Free Dense Structure from Motion (\ours{}). Rather than forcing all observations of a track onto one 3D point, we lift both pixels of each correspondence to separate 3D Gaussians, which align by moving their centers together or, for unreliable matches, by growing their variances. This volumetric uncertainty smooths the highly non-convex landscape and enables a strictly \textbf{initialization-free} regime, in the precise sense that the front-end supplies only 2D correspondences and a covisibility topology (carrying no metric pose, depth, or calibration), while the geometric optimization starts cold from identity poses, uniform depths, and uninformed intrinsics. We make no global-optimality claim. This cold-start convergence is \emph{empirical}, demonstrated on a controlled toy problem (\Cref{sec:objective}) and across all four benchmarks (\Cref{sec:experiments}). As shown in Figure~\ref{fig:teaser}, \ours{} recovers dense 3D geometry across diverse, in-the-wild scenarios that are challenging for classical baselines.

\medskip\noindent In summary, our core contributions are:
\begin{itemize}
    \item \textbf{Probabilistic landmarks and a unified NLL objective.} We replace point tracks with per-observation 3D Gaussians under a Negative Log-Likelihood (NLL) objective over 2D reprojection and 3D consistency. Optimizing each correspondence's variance down-weights noisy matches and broadens the basin of attraction, enabling initialization-free optimization (\Cref{sec:objective}).
    \item \rev[b1]{\textbf{Relative-pose parameterization over a spanning tree}, as in MASt3R-SfM~\cite{duisterhof2025mast3rsfm}. We show that it is what keeps the cold start stable (\Cref{tab:ablation}), and we compute the chaining with an $O(\log N)$ parallel prefix scan rather than a serial traversal.}
    \item \textbf{Dual-hypothesis mirror resolution and adaptive edge weighting}: the former resolves the mirror-symmetry ambiguity, the latter improves accuracy by down-weighting unreliable view-graph edges (\Cref{sec:mirror_reg,sec:adaptedge}).
\end{itemize}

\section{Related Work}
\label{sec:related}

\paragraph{Relative Pose Estimation and Graph Optimization.} 
The estimation of camera poses from sparse correspondences is a foundational problem in computer vision~\cite{hartley2003multiple, ozyecsil2017survey}. Standard pipelines like \textit{COLMAP}~\cite{schonberger2016sfm} rely on the incremental construction of multi-view tracks, which are computationally expensive and highly sensitive to noise when scaled to dense correspondences. \rev[b3]{\textit{GLOMAP}~\cite{pan2024glomap} solves rotation averaging and then global positioning over the entire view graph, which removes the ordering dependence that makes incremental SfM brittle. Its global rotations are nonetheless averaged from pairwise relative rotations estimated on the same correspondences, and are held fixed during positioning, so outliers among noisy dense matches corrupt the geometry that the final bundle adjustment starts from.} While methods like the \textit{detector-free Structure-from-Motion} pipeline~\cite{he2024dsfm} attempt to mitigate this by constructing robust feature tracks to adapt dense matches into traditional SfM, enforcing rigid multi-view consistency remains a bottleneck. Among recent dense matchers~\cite{edstedt2023dkm, edstedt2024roma}, we use Dense Kernelized feature Matching (\textit{DKM})~\cite{edstedt2023dkm} to extract dense pairwise correspondences. Rather than forcing these dense matches into rigid global tracks, \ours{} performs joint optimization directly over a relative-pose graph, dynamically down-weighting inconsistent edges~\cite{sunderhauf2012switchable, yang2020graduated}.

\paragraph{Probabilistic Scene Representations.} 
Modeling spatial uncertainty is critical for robust reasoning in unstructured environments. While EKF-based SLAM systems~\cite{davison2007monoslam} track covariances, they often suffer from linearization errors. Probabilistic treatments of BA assume Gaussian observation noise and propagate it to the estimates~\cite{kaess2012isam2, Kyle2020uncervis}, yet these methods remain largely point-based, ignoring the volumetric uncertainty of 3D landmarks. A key distinction of our work is shifting from these point-based models to volumetric 3D Gaussians, which propagate uncertainty from the 2D image plane into the 3D domain and yield a consistency loss that smooths the landscape toward prior-free convergence.

\paragraph{Initialization-Free Bundle Adjustment.} 
Recent research has focused on expanding the basin of attraction for BA to enable convergence from identity or random initializations. Hong~\etal{} addressed this with the variable-projection (\textit{VarPro}) method~\cite{Hong2016projba}, which analytically eliminates the 3D structure and optimizes over cameras alone, and with the pseudo-object space error (\textit{pOSE})~\cite{Hong_2018_CVPR}, an object-space surrogate that suppresses trivial solutions. This line was scaled by a power variable projection (\textit{PoVar})~\cite{weber2024power} and extended with an exponential depth regularization (\textit{expOSE})~\cite{iglesias2023expose}, while \textit{Building Rome with Convex Optimization}~\cite{han2025building} uses convex relaxations with a learned-depth prior. These methods broaden the basin while still solving for a single \emph{exact, shared} 3D point per track, a rigid formulation tailored to sparse, outlier-free correspondences that becomes brittle under the noisy dense matches of modern matchers. \ours{} instead relaxes the shared point into a per-observation probabilistic 3D Gaussian, makes each correspondence's variance a \emph{free, jointly optimized} variable rather than a fixed surrogate weight, and resolves the mirror-symmetry ambiguity via dual-hypothesis regularization, all from a strict cold start.

\paragraph{Deep-Learning-Based SfM.}
A parallel line of research bypasses classical SfM, either by optimizing poses through a differentiable renderer or by deep-learning-based regression. Frameworks such as \textit{COLMAP-Free 3D Gaussian Splatting}~\cite{fu2024colmapfree} reconstruct without SfM preprocessing but often rely on sequential, video-like inputs. Furthermore, ``all-in-one''  foundation models like \textit{DUSt3R}~\cite{wang2024dust3r} and \textit{MASt3R}~\cite{revaud2024mast3r} directly regress 3D point maps from image pairs, while subsequent extensions like \textit{MASt3R-SfM}~\cite{duisterhof2025mast3rsfm} adapt these pairwise predictions for global Structure-from-Motion. Similarly, differentiable pipelines like \textit{VGGSfM}~\cite{wang2024vggsfm} and \textit{VGGT}~\cite{vggt2024} extract global camera geometry without relying on classical multi-view tracks. Although these methods are robust in unconstrained scenarios, they function largely as black-box predictors. \mbox{\ours{}} shares their freedom from initialization but remains an explicit geometric optimization over image-pair correspondences, so each result can be traced back to individual matches and their variances. %
\newcommand{\se}{\mathfrak{se}}
\newcommand{\SE}{\mathbf{SE}}

\section{Methodology}
\label{sec:method}

We present \ours{}, a framework that replaces the rigid tracks of classical pipelines with 3D Gaussians and jointly optimizes camera extrinsics, focal lengths, and scene geometry without any prior metric pose or intrinsic initialization.

\medskip\noindent\textbf{Framework Overview.} The core idea behind \ours{} is to relax the rigid \emph{point track} into a probabilistic 3D landmark under a Negative Log-Likelihood (NLL) objective, which broadens the cold-start basin. We test this on a minimal example (\Cref{sec:objective}) and quantify it by ablation (\Cref{sec:ablation}). On top of this relaxation, optimizing two mirrored hypotheses in parallel resolves the mirror-symmetry ambiguity on object-centric scenes (\Cref{sec:mirror_reg}). Adaptive edge weighting then improves accuracy by down-weighting unreliable view-graph edges (\Cref{sec:adaptedge}). As illustrated in Figure~\ref{fig:overview}, \ours{} operates over a view graph $\mathcal{G} = \{V, E\}$ and its spanning tree $\mathcal{T}$, where $V$ holds the $N$ cameras (one per image) and $E$ the image pairs that are densely matched. To build the graph, we first run LoFTR~\cite{sun2021loftr} at a coarse resolution on every image pair to measure how much the two images overlap (their covisibility). We then apply the expensive \emph{dense} matcher DKM~\cite{edstedt2023dkm} only to a sparse set of high-overlap pairs. \rev[b6]{Appendix~\ref{sec:viewgraph} (Algorithm~\ref{alg:mst}) details how the tree is built and how each image then gains extra edges to its $N_{\text{neig}}{-}1$ most covisible images. The reconstruction is robust to the choice of tree (\Cref{sec:mst_robustness}).} The tree also serves as a kinematic chain: each camera pose is parameterized relative to its parent, which stabilizes the cold start (\Cref{sec:kinematic}).

\subsection{Preliminary: Kinematic Pose Parameterization}
\label{sec:kinematic}

To stabilize prior-free optimization, we parameterize the camera poses kinematically over the spanning tree $\mathcal{T}$, \rev[b1kin]{similar to MASt3R-SfM~\cite{duisterhof2025mast3rsfm}}. We write $\mathbf{T}_{AB}$ for the rigid transform mapping points from frame $B$ to frame $A$, so $\mathbf{T}_{wi}$ is the camera-to-world pose and $\mathbf{T}_{ij}$ the relative pose from frame $j$ to frame $i$. We designate a root camera $r$ as the world origin by fixing its pose to the identity matrix ($\mathbf{T}_{rr} = \mathbf{I}$). For all other cameras $i \in V \setminus \{r\}$, we optimize their pose strictly relative to their parent node $P(i)$ in the tree.

\medskip\noindent\textbf{Relative Transformation Parametrization.} We parameterize the relative transformation using a minimal 6-DOF vector $\boldsymbol{\delta}_i \in \mathbb{R}^6$. This vector is mapped to the Lie algebra $\mathfrak{se}(3)$ via the hat operator $\hat{(\cdot)}$, and subsequently projected to the Lie group $\mathbf{SE}(3)$ using the exponential map: $\mathbf{T}_{P(i), i} = \exp(\hat{\boldsymbol{\delta}}_i)$. This tree-based kinematic chain naturally propagates parent updates to all descendants, smoothing the optimization landscape and steering it away from poor local minima. \rev[b1ev]{\Cref{fig:toy} isolates this effect on a controlled instance and \Cref{sec:ablation} quantifies it on real scenes.}

The absolute pose $\mathbf{T}_{ri}$ is obtained by chaining relative transformations along the path $\mathcal{P}(r \to i)$:
\begin{equation}
    \mathbf{T}_{ri} = \prod_{j \in \mathcal{P}(r \to i) \setminus \{r\}} \exp(\hat{\boldsymbol{\delta}}_j)
\end{equation}
Furthermore, the sequential chaining introduces an $O(N)$ bottleneck. For scalability, we compute the chaining via a parallel prefix scan~\cite{harris2007parallel} in $O(\log N)$ steps.

\subsection{Gaussian Landmark Relaxation}
\label{sec:objective}

Classical bundle adjustment represents each track by one shared 3D point under a quadratic reprojection cost. From a strict cold start (identity cameras, no metric depths), it stalls near its initialization. \ours{} instead models each landmark as an isotropic 3D Gaussian. The key difference is that the per-correspondence variance is itself \emph{optimized} rather than fixed. An unreliable match thus inflates its variance and is down-weighted. Large variances also smooth the objective, which broadens the basin of attraction.

\medskip\noindent\textbf{Probabilistic model.} Consider a single correspondence: pixel $\mathbf{u}$ in a source view $i$ matched to pixel $\mathbf{v}$ in a target view $j$. Pixel-level quantities (depth $d_u$, variance $\sigma_u^2$, point $\mathbf{p}_u$) are subscripted by pixel, and per-view ones (intrinsics $\mathbf{K}_i$, pose $\mathbf{T}_{ri}$) by frame. Each pixel has its own 3D landmark $\mathbf{X}_u$: the back-projected point $\mathbf{p}_u = \pi^{-1}(\mathbf{u}, d_u, \mathbf{K}_i)$, obtained by lifting pixel $\mathbf{u}$ to depth $d_u$ in camera $i$'s frame, is modeled as $\mathbf{p}_u = \mathbf{X}_u + \boldsymbol{\varepsilon}_u$ with $\boldsymbol{\varepsilon}_u \sim \mathcal{N}(\mathbf{0}, \sigma_u^2 \mathbf{I})$. Minimizing the sum of the two NLL terms below is joint maximum-likelihood estimation of the geometry and the per-correspondence noise scales, and their log-variance terms penalize the trivial infinite-variance solution.

Concretely, the landmark is the isotropic Gaussian $\mathcal{G}(\mathbf{p}_u, \sigma_u^2 \mathbf{I}_{3\times 3})$ centered at $\mathbf{p}_u$. The depth $d_u$ and variance $\sigma_u^2$ are sampled at pixel $\mathbf{u}$ from view $i$'s optimizable $32 \times 32$ depth and variance maps (to reduce the number of parameters), while the intrinsics $\mathbf{K}_i$ are shared across all of view $i$'s observations. We keep the covariance isotropic because anisotropic Gaussians overfit the inaccurate early poses and destabilize the joint optimization (\Cref{sec:iso_aniso}).

\medskip\noindent\textbf{2D Reprojection Objective.}
We transform the source point into the target frame, $\mathbf{p}_{u \to j} = (\mathbf{T}_{jr} \mathbf{T}_{ri}) \mathbf{p}_u$, and project the transformed Gaussian onto the target image plane, yielding covariance $\boldsymbol{\Sigma}_{2D} = \sigma_u^2 \, \mathbf{J} \mathbf{J}^\top$, where $\mathbf{J} = \partial \pi / \partial \mathbf{p}_{u \to j} \in \mathbb{R}^{2 \times 3}$ is the Jacobian of the projection $\pi$ at $\mathbf{p}_{u \to j}$. The 2D reprojection loss is the Negative Log-Likelihood (NLL) of the target pixel $\mathbf{v}$, with residual $\mathbf{r} = \pi(\mathbf{p}_{u \to j}) - \mathbf{v}$:
\begin{equation}
    \mathcal{L}_{2D}^{i \to j} = \tfrac{1}{2}\mathbf{r}^{\top} \boldsymbol{\Sigma}_{2D}^{-1}\mathbf{r} + \tfrac{1}{2}\log\det\boldsymbol{\Sigma}_{2D} + \log(2\pi).
\end{equation}

\medskip\noindent\textbf{Soft Spatial Constraint (SSC).}
Since the two pixels should see the same scene point, their landmarks $\mathbf{X}_u$ and $\mathbf{X}_v$ should coincide in 3D. Rather than forcing them onto an exact shared coordinate, we softly maximize the probabilistic overlap of their Gaussians: corresponding Gaussians align either by shifting their centers together or by growing their variance, the latter absorbing outliers and inaccurate poses\rev{, which we quantify by sweeping corruption ratio against noise magnitude in \Cref{sec:noise_robustness}}. The overlap of $\mathcal{G}(\mathbf{p}_{u \to j}, \sigma_{u}^{2}\mathbf{I}_{3\times 3})$ and $\mathcal{G}(\mathbf{p}_{v}, \sigma_{v}^{2}\mathbf{I}_{3\times 3})$ is a Gaussian in the residual $\mathbf{r} = \mathbf{p}_{u \to j} - \mathbf{p}_{v}$ with covariance $\boldsymbol{\Sigma}_{3D} = (\sigma_{u}^{2} + \sigma_{v}^{2})\mathbf{I}_{3\times 3}$, so the 3D consistency loss takes the same NLL form:
\begin{equation}
    \mathcal{L}_{3D}^{ij} = \tfrac{1}{2}\mathbf{r}^{\top} \boldsymbol{\Sigma}_{3D}^{-1}\mathbf{r} + \tfrac{1}{2}\log\det\boldsymbol{\Sigma}_{3D} + \tfrac{3}{2}\log(2\pi).
\end{equation}

\medskip\noindent\textbf{Unified Landmark Objective.}
For each edge $e_{ij}$, we sum both reprojection directions and the 3D term over all of its correspondences. This per-edge NLL is the core loss function of \ours{}:
\begin{equation}\label{eq:unified}
    \boxed{\;\mathcal{L}^{ij} = \sum_{\text{matches}} \left( \mathcal{L}_{2D}^{i \to j} + \mathcal{L}_{2D}^{j \to i} + \mathcal{L}_{3D}^{ij} \right)\;}
\end{equation}

\begin{figure}[t]
    \centering
    \includegraphics[height=0.46\linewidth]{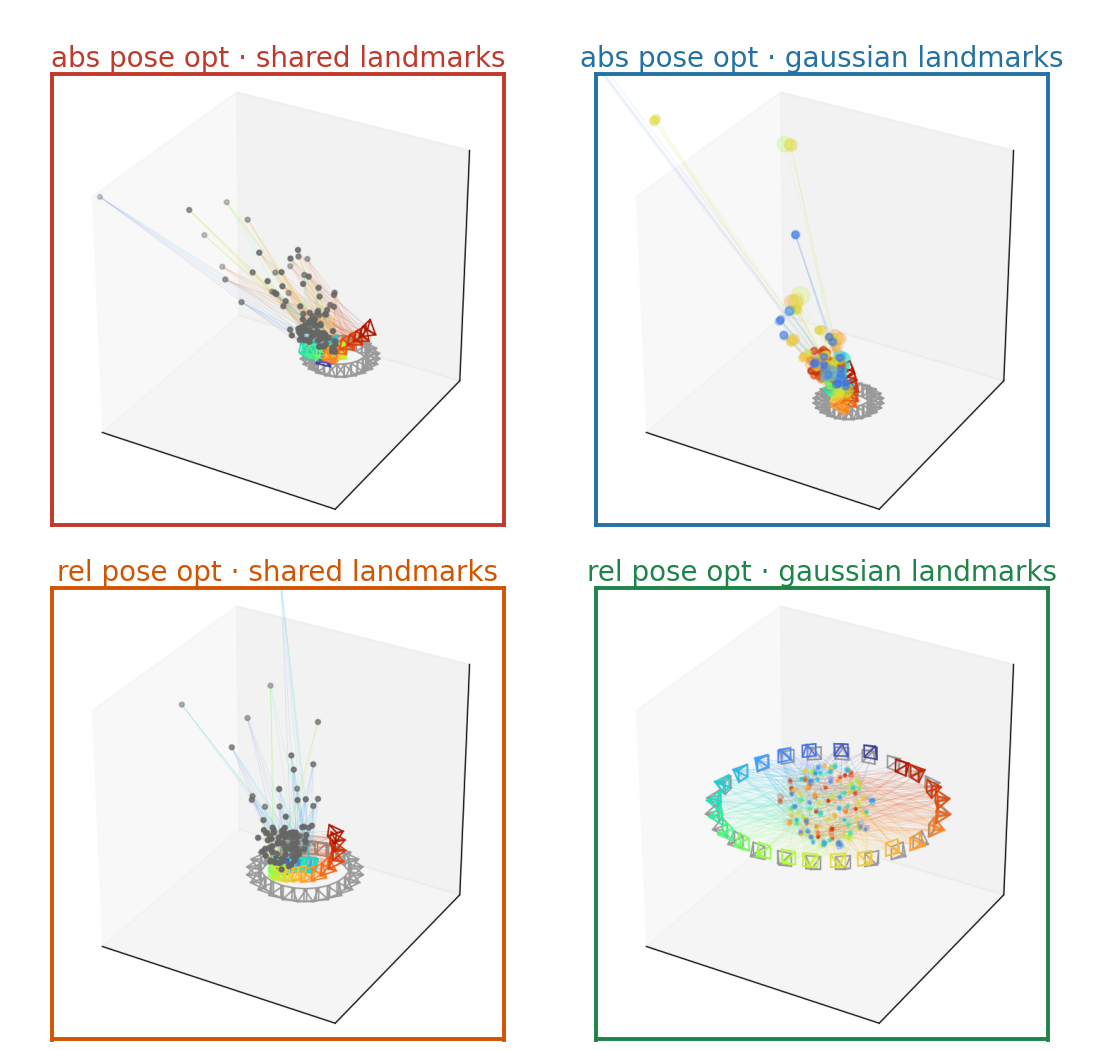}\hfill
    \includegraphics[height=0.46\linewidth]{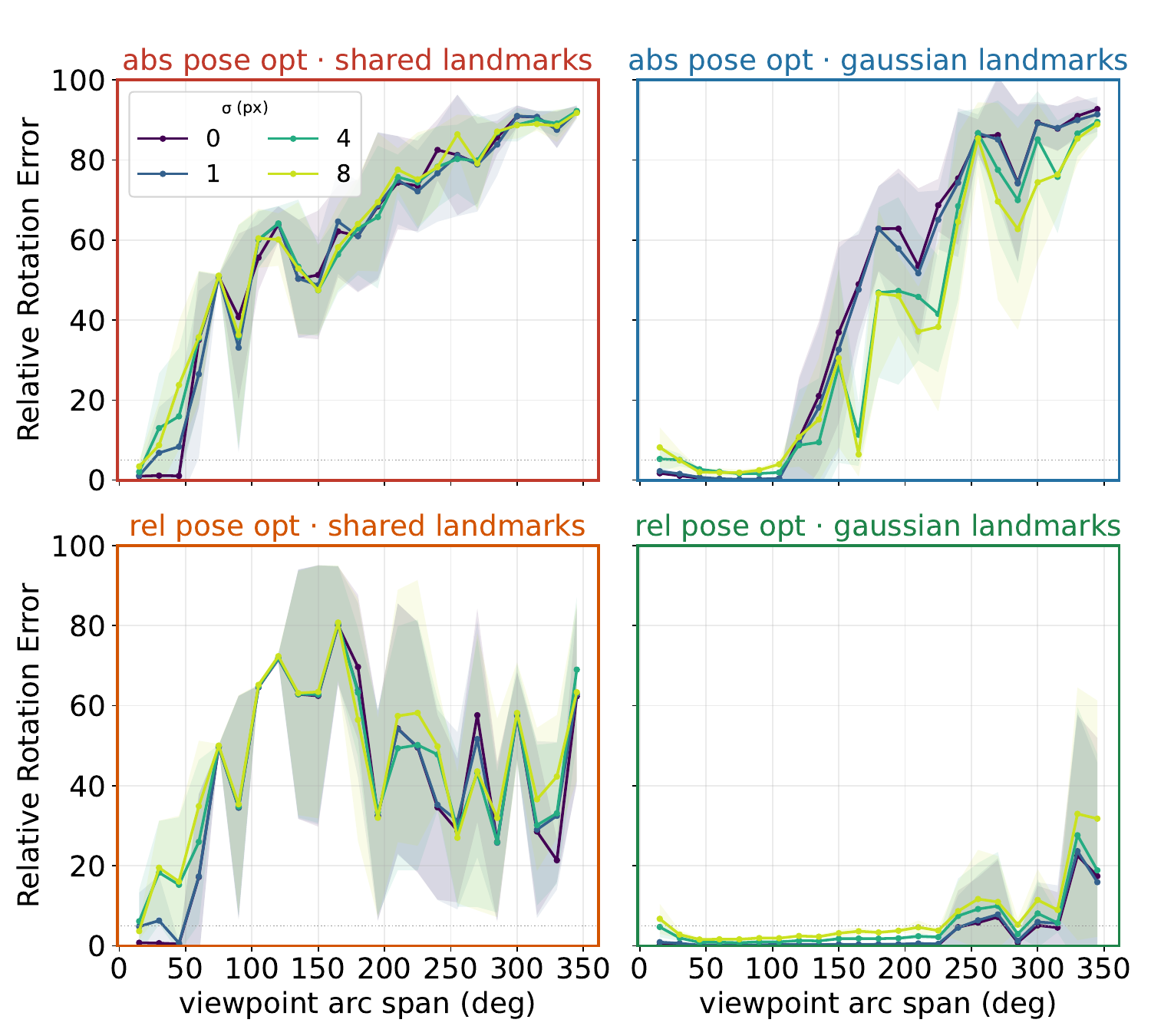}
    \caption{\textbf{Cold-start convergence: toy ablation.} Pose parametrization (independent abs-poses vs.\ chained relative-pose) $\times$ landmark model (one shared 3D point, i.e.\ classical BA, vs.\ per-observation 3D Gaussians); \emph{Left:} reconstructions on the full $360^\circ$ ring ($24$ cameras, $\sigma{=}4$\,px). \emph{Right:} mean pairwise relative-rotation error vs.\ arc span ($+15^\circ$ per added camera), over noise $\sigma\!\in\!\{0,1,4,8\}$\,px and $5$ seeds. Only chained poses \emph{and} Gaussian landmarks (green) recover the ring; all other cells collapse.}
    \label{fig:toy}
\end{figure}

\medskip\noindent\textbf{Toy Model.}
Before applying \ours{} to real image sets, we first establish a proof of concept, that the objective (\Cref{eq:unified}) is able to converge from a strict cold start, on a controlled instance (\Cref{fig:toy}): cameras on a widening arc viewing a sphere of landmarks, initialized from identity poses and unit depth (full details in \Cref{sec:toy_setup}).

We claim no global optimality, but observe an empirically broad \emph{basin of attraction}, and that \emph{both} ingredients are necessary. With Gaussian landmarks \emph{and} the relative-pose chain, the cameras converge from a cold start (identity poses, uniform depth) to their ground-truth poses across the full $360^\circ$ ring, and the landmarks to the correct 3D structure. Removing either ingredient fails: \revlabel{b2toy}the use of independent absolute poses degrades sharply beyond $120^\circ$, and classical BA stays trapped near its initialization at every viewpoint span and noise level.

\subsection{Robustifying the Optimization}
\label{sec:robust}
On top of this relaxation, we first resolve the mirror-symmetry ambiguity that arises on object-centric scenes, then push accuracy higher by reweighting the view graph.

\subsubsection{Dual-Hypothesis Mirror Resolution}
\label{sec:mirror_reg}

\begin{figure}[H]
    \centering
    \includegraphics[width=1.0\linewidth]{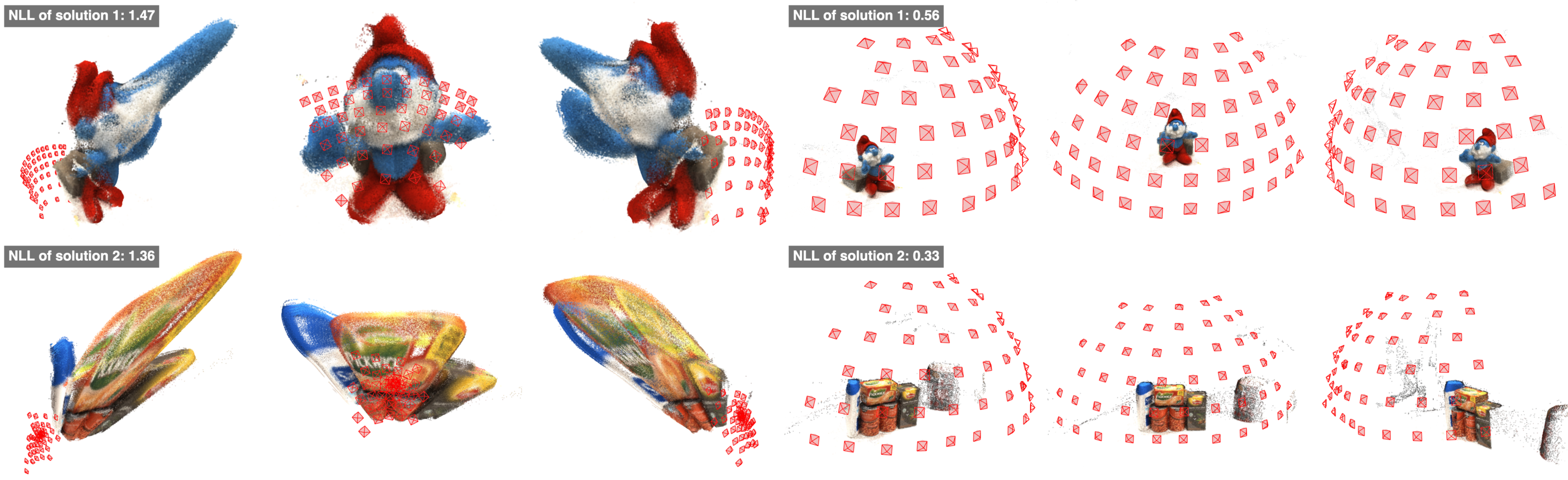}
    \caption{\textbf{The mirror-symmetry ambiguity.} A scene and its reflection (concave surfaces appear convex) render near-identically, so the optimizer can fall into the reflected basin; \ours{} optimizes both hypotheses and keeps the lower-NLL one.}
    \label{fig:mirror_ambiguity}
    \vspace{-0.4cm}
\end{figure}

Our SE(3) and pinhole parameterization restricts the search to metric reconstructions, ruling out general projective ambiguities~\cite{hartley2003multiple}. The depth-sign ambiguity is the remaining discrete trap on low-parallax object-centric scenes. This \emph{mirror-symmetry} ambiguity~\cite{cheng2023lu, ozden2010multibody} arises because, when perspective effects are weak, the projection is nearly affine: a scene and its mirror image, reflected across a fronto-parallel plane so that depth inverts and a concave surface becomes convex, project to near-identical images, so structure and motion admit two near-indistinguishable solutions (\Cref{fig:mirror_ambiguity}).

This ambiguity is a \emph{discrete} basin that gradient descent falls into from a cold start, which global alignment does not resolve, and empirically the dominant failure mode on object-centric scenes (\Cref{sec:ablation}). We resolve it without metric priors by concurrently optimizing two pose hypotheses: a primary world and a secondary one seeded into the reflected basin.

On the camera side, the two solutions are related by a $180^\circ$ rotation about the optical axis, $\mathbf{M} = \operatorname{diag}(-1,-1,1,1)$. Despite the name ``mirror'', $\mathbf{M}$ is a \emph{proper} rotation ($\det\mathbf{M}=+1$) along the optical axis, not a spatial reflection: the reflection shows up only in the recovered 3D structure (an inverted disparity map). Conjugating each relative pose, $\mathbf{T}^{\text{mir}}_{ij} = \mathbf{M} \, \mathbf{T}_{ij} \, \mathbf{M}$, reorients the cameras into the reflected configuration, equivalently pre-multiplying the world-to-camera rotations by this optical-axis rotation, as in unposed NeRF~\cite{cheng2023lu}. We seed the secondary hypothesis with $\mathbf{T}^{\text{mir}}_{ij}$ and, during early optimization, regularize its relative poses $\boldsymbol{\xi}^{\text{mir}}_{ij} \in \mathbb{R}^6$ toward it:
\begin{equation}
    \label{eq:reg_loss}
    \boxed{\;\mathcal{L}_{\text{reg}} = \lambda_{\text{mir}}(t) \; \frac{1}{|E|}\sum_{e_{ij} \in E} \left\| \text{sg}\left(\log\!\left(\mathbf{T}^{\text{mir}}_{ij}\right)\right) - \boldsymbol{\xi}^{\text{mir}}_{ij} \right\|_1\;}
\end{equation}
where $\lambda_{\text{mir}}(t)$ anneals to zero over an early warm-up (schedule in \Cref{sec:dual_hypothesis_dynamics}).

The stop-gradient ($\text{sg}$) steers the secondary hypothesis toward a valid reflection without distorting the primary solution. This regularization prevents the two worlds from collapsing into a single trivial state, while the annealing permits the later unconstrained refinement of the secondary hypothesis.

\subsubsection{Adaptive Edge Weighting}
\label{sec:adaptedge}

We attach a reliability weight to each edge to down-weight unreliable links, letting geometrically valid connections dominate the joint optimization (validated in \Cref{sec:ablation}).

We can weight each edge's contribution to each world's objective $\mathcal{L}_{\text{world}}$ by a reliability $W_{ij}^{(t)} \in [0,1]$, high for edges that fit well relative to their neighbors at a shared camera and low for outliers, derived from the local NLL statistics and smoothed over time (the $z$-score, logistic map, and EMA schedule are deferred to \Cref{sec:impl}). It is held constant during backpropagation via a stop-gradient ($\text{sg}$). Otherwise, the optimizer could lower the loss simply by shrinking the weights, silencing hard edges, rather than improving the geometry. The weighted per-world objective is thus:
\begin{equation}\label{eq:world_loss}
    \boxed{\;\mathcal{L}_{\text{world}} = \sum_{e_{ij} \in E} \text{sg}\left(W_{ij}^{(t)}\right) \mathcal{L}^{ij}\;}
\end{equation}

\medskip\noindent\textbf{Final Training Objective.} Let $\mathcal{L}_{\text{world}}^{(1)}$ and $\mathcal{L}_{\text{world}}^{(2)}$ denote this weighted per-world objective (\Cref{eq:world_loss}) for the primary and secondary hypotheses, which maintain completely independent sets of learnable parameters (separate camera extrinsics, intrinsics, and 3D Gaussian attributes). The complete joint objective combines both worlds with the mirror regularizer of \Cref{eq:reg_loss}:
\begin{equation}
\label{eq:final_loss}
    \boxed{\;\mathcal{L}_{\text{final}} = \mathcal{L}_{\text{world}}^{(1)} + \mathcal{L}_{\text{world}}^{(2)} + \mathcal{L}_{\text{reg}}\;}
\end{equation}

Minimizing this joint objective allows \ours{} to concurrently explore both worlds without parameter entanglement. Parallel optimization ensures our reported runtimes encompass this dual exploration. Upon convergence, selecting the world with the lower $\mathcal{L}_{\text{world}}$ effectively bypasses the local minimum of the mirrored solution (\Cref{fig:dual_world_loss}). %

\section{Experiments}
\label{sec:experiments}

\begin{table*}[t]
\centering
\caption{\textbf{Aggregated Metrics Comparison on DTU, Mip-NeRF 360, ETH3D, and IMC2021.} Comparison of Relative Rotation Accuracy (RRA), Relative Translation Accuracy (RTA) \rev{and pose AUC (\Cref{sec:metrics})}. Within each group, the \textbf{best} and \underline{second-best} per column are marked in bold and underline; cell shading runs from green (high) to red (low). \rev{Average images per scene: DTU 49, Mip-NeRF~360 232, ETH3D 35; IMC2021 evaluates 5/10/25-view bags.} Note that IMC2021 uses different thresholds (@3\textdegree, @5\textdegree, @10\textdegree) due to its in-the-wild nature.}
\label{tab:aggregated_results_all_horizontal}
\begin{adjustbox}{max width=\linewidth}
\begin{tabular}{l | ccc | ccc | c | ccc | ccc | c | ccc | ccc | c | ccc | ccc | c}
\toprule
\multirow{3}{*}{Methods} & \multicolumn{7}{c|}{DTU\cite{Jensen_2014_CVPR}} & \multicolumn{7}{c|}{Mip-NeRF 360\cite{barron2022mipnerf360}} & \multicolumn{7}{c|}{ETH3D\cite{schops2017multi}} & \multicolumn{7}{c}{IMC2021\cite{jin2021image}} \\
\cmidrule(lr){2-8} \cmidrule(lr){9-15} \cmidrule(lr){16-22} \cmidrule(l){23-29}
 & \multicolumn{3}{c|}{RRA} & \multicolumn{3}{c|}{RTA} & \multicolumn{1}{c|}{\rev{AUC}} & \multicolumn{3}{c|}{RRA} & \multicolumn{3}{c|}{RTA} & \multicolumn{1}{c|}{\rev{AUC}} & \multicolumn{3}{c|}{RRA} & \multicolumn{3}{c|}{RTA} & \multicolumn{1}{c|}{\rev{AUC}} & \multicolumn{3}{c|}{RRA} & \multicolumn{3}{c|}{RTA} & \multicolumn{1}{c}{\rev{AUC}} \\
 & @1$^\circ$ & @5$^\circ$ & @10$^\circ$ & @1$^\circ$ & @5$^\circ$ & @10$^\circ$ & \rev{@10$^\circ$} & @1$^\circ$ & @5$^\circ$ & @10$^\circ$ & @1$^\circ$ & @5$^\circ$ & @10$^\circ$ & \rev{@10$^\circ$} & @1$^\circ$ & @5$^\circ$ & @10$^\circ$ & @1$^\circ$ & @5$^\circ$ & @10$^\circ$ & \rev{@10$^\circ$} & @3$^\circ$ & @5$^\circ$ & @10$^\circ$ & @3$^\circ$ & @5$^\circ$ & @10$^\circ$ & \rev{@10$^\circ$} \\
\midrule
\multicolumn{25}{l}{\textit{With Deep Scene Prior}} \\
MASt3R-SfM\cite{duisterhof2025mast3rsfm} & \hcu{61.2} & \hcu{99.6} & \hcb{100.0} & \hcu{57.3} & \hcu{98.1} & \hcu{99.7} & \rev{\hcu{85.1}} & \rev{\hcu{29.2}} & \rev{\hcu{87.1}} & \rev{\hcu{91.9}} & \rev{\hcu{35.0}} & \rev{\hcu{82.2}} & \rev{\hcu{89.5}} & \rev{\hcu{67.0}} & \rev{\hcb{63.5}} & \rev{\hcu{78.9}} & \rev{\hcu{83.2}} & \rev{\hcb{54.0}} & \rev{\hcu{74.3}} & \rev{\hcu{79.5}} & \rev{\hcb{66.5}} & \hcu{57.5} & \hcu{75.1} & \hcu{83.4} & \hcu{52.8} & \hcu{69.2} & \hcu{81.6} & \rev{\hcu{67.1}} \\
VGGT\cite{vggt2024} & \hcb{72.4} & \hcb{100.0} & \hcb{100.0} & \hcb{75.8} & \hcb{100.0} & \hcb{100.0} & \rev{\hcb{90.8}} & \hcbr{62.2} & \hcbr{99.9} & \hcbr{100.0} & \hcbr{50.6} & \hcbr{95.1} & \hcbr{98.3} & \hcbr{82.4} & \hcu{22.3} & \hcb{96.2} & \hcb{98.6} & \hcu{22.6} & \hcb{81.1} & \hcb{92.2} & \rev{\hcu{53.9}} & \rev{\hcb{99.7}} & \rev{\hcb{100.0}} & \rev{\hcb{100.0}} & \rev{\hcb{69.1}} & \rev{\hcb{81.1}} & \rev{\hcb{90.9}} & \rev{\hcb{72.6}} \\
\midrule
\multicolumn{25}{l}{\textit{Without Deep Scene Prior}} \\
COLMAP\cite{schonberger2016sfm} & \hcr{74.9} & \hcr{87.3} & \hcr{87.3} & \hcr{36.1} & \hcr{74.6} & \hcr{82.6} & \hcr{65.8} & \hcbr{100.0} & \hcb{100.0} & \hcb{100.0} & \hcbr{98.2} & \hcb{100.0} & \hcb{100.0} & \hcbr{97.9} & \hcbr{90.5} & \hcur{90.6} & \hcur{90.6} & \hcbr{80.1} & \hcur{89.2} & \hcur{90.1} & \hcur{84.8} & \rev{\hc{81.9}} & \rev{\hc{85.3}} & \rev{\hc{88.4}} & \rev{\hc{54.9}} & \rev{\hc{65.3}} & \rev{\hc{75.4}} & \rev{\hc{58.9}} \\
COLMAP + DKM\cite{edstedt2023dkm} & \hcur{88.5} & \hcur{97.1} & \hcur{97.2} & \hcur{62.2} & \hcbr{93.2} & \hcbr{95.9} & \hcbr{83.7} & \hcr{79.1} & \hcr{79.1} & \hcr{79.1} & \hcr{78.1} & \hcr{79.1} & \hcr{79.1} & \hcr{77.6} & \hcr{46.6} & \hcr{77.1} & \hcr{80.3} & \hcr{23.2} & \hcr{59.3} & \hcr{69.1} & \hcr{51.0} & \rev{\hcu{91.1}} & \rev{\hcu{94.3}} & \rev{\hcu{96.5}} & \rev{\hcu{63.3}} & \rev{\hcu{75.3}} & \rev{\hcu{86.6}} & \rev{\hcu{67.4}} \\
COLMAP + RoMa\cite{edstedt2024roma} & \hcbr{89.5} & \hcr{96.6} & \hcr{96.8} & \hcbr{62.8} & \hcbr{93.2} & \hcur{95.8} & \hcbr{83.7} & \hcr{83.2} & \hcr{83.2} & \hcur{83.2} & \hcr{82.2} & \hcr{83.2} & \hcr{83.2} & \hcr{81.6} & \hcr{52.6} & \hcr{72.6} & \hcr{79.7} & \hcr{28.8} & \hcr{60.1} & \hcr{68.3} & \hcr{52.0} & \rev{\hcb{92.4}} & \rev{\hcb{95.1}} & \rev{\hcb{97.2}} & \rev{\hcb{66.4}} & \rev{\hcb{78.0}} & \rev{\hcb{87.9}} & \rev{\hcb{69.7}} \\
pOSE\cite{Hong_2018_CVPR} & \hc{50.1} & \hc{64.0} & \hc{67.0} & \hc{51.8} & \hc{65.1} & \hc{68.2} & \rev{\hc{59.7}} & \hc{0.2} & \hc{3.3} & \hc{6.8} & \hc{0.1} & \hc{2.1} & \hc{6.2} & \rev{\hc{0.9}} & \hc{0.1} & \hc{3.7} & \hc{10.0} & \hc{0.2} & \hc{2.8} & \hc{7.5} & \rev{\hc{0.9}} & \hc{41.0} & \hc{49.9} & \hc{63.5} & \hc{25.0} & \hc{31.7} & \hc{43.6} & \rev{\hc{28.4}} \\
PoVar\cite{weber2024power} & \hc{57.0} & \hc{69.7} & \hc{72.3} & \hc{58.9} & \hc{69.9} & \hc{73.0} & \rev{\hc{64.8}} & \hc{0.3} & \hc{4.8} & \hc{10.0} & \hc{0.1} & \hc{2.4} & \hc{7.7} & \rev{\hc{0.9}} & \hc{0.1} & \hc{2.7} & \hc{7.1} & \hc{0.1} & \hc{1.0} & \hc{3.8} & \rev{\hc{0.4}} & \hc{38.3} & \hc{48.6} & \hc{61.3} & \hc{20.6} & \hc{28.0} & \hc{40.7} & \rev{\hc{23.9}} \\
pOSE + DKM\cite{edstedt2023dkm} & \hc{8.8} & \hc{13.6} & \hc{17.2} & \hc{5.6} & \hc{14.6} & \hc{18.6} & \rev{\hc{11.1}} & \hc{0.2} & \hc{2.5} & \hc{4.3} & \hc{0.1} & \hc{1.6} & \hc{3.9} & \rev{\hc{0.8}} & \hc{2.5} & \hc{6.8} & \hc{13.3} & \hc{0.6} & \hc{4.3} & \hc{10.5} & \rev{\hc{2.8}} & \hc{43.7} & \hc{55.3} & \hc{69.0} & \hc{21.8} & \hc{29.8} & \hc{41.5} & \rev{\hc{26.1}} \\
PoVar + DKM\cite{edstedt2023dkm} & \hc{10.7} & \hc{20.2} & \hc{24.4} & \hc{6.1} & \hc{20.9} & \hc{26.7} & \rev{\hc{15.8}} & \hc{0.2} & \hc{4.4} & \hc{6.3} & \hc{0.1} & \hc{1.8} & \hc{5.9} & \rev{\hc{0.8}} & \hc{2.7} & \hc{6.0} & \hc{12.6} & \hc{1.0} & \hc{2.0} & \hc{6.3} & \rev{\hc{1.6}} & \hc{32.8} & \hc{42.3} & \hc{57.5} & \hc{14.9} & \hc{20.3} & \hc{32.1} & \rev{\hc{18.3}} \\
\rev{\textbf{\ours{}}} & \rev{\hc{88.1}} & \rev{\hcb{97.9}} & \rev{\hcb{99.8}} & \rev{\hc{52.2}} & \rev{\hcu{83.1}} & \rev{\hc{91.7}} & \rev{\hcu{75.0}} & \rev{\hcu{96.7}} & \rev{\hcu{99.6}} & \rev{\hcb{100.0}} & \rev{\hcu{92.2}} & \rev{\hcu{98.5}} & \rev{\hcu{99.2}} & \rev{\hcu{94.0}} & \rev{\hcu{86.3}} & \rev{\hcb{96.9}} & \rev{\hcb{99.0}} & \rev{\hcu{77.4}} & \rev{\hcb{90.1}} & \rev{\hcb{92.4}} & \rev{\hcb{85.0}} & \rev{\hc{59.5}} & \rev{\hc{71.2}} & \rev{\hc{84.0}} & \rev{\hc{29.9}} & \rev{\hc{39.5}} & \rev{\hc{52.4}} & \rev{\hc{37.1}} \\
\bottomrule
\end{tabular}
\end{adjustbox}
\end{table*}

\begin{table*}[t]
\centering
\caption{\textbf{Per-Scene Metrics Comparison on ETH3D.} Comparison of Relative Rotation Accuracy (RRA@5$^\circ$), Relative Translation Accuracy (RTA@5$^\circ$), Accuracy at 5cm (ACC@5cm), Completeness at 5cm (COMP@5cm) \rev{and pose AUC@5$^\circ$} across different baselines. Within each row, the \textbf{best} per metric is marked in bold. Parentheses give the number of images per scene.}
\label{tab:per_scene_metrics_compact}
\begin{adjustbox}{max width=\linewidth}
\setlength{\tabcolsep}{3pt} 
\begin{tabular}{l | ccccc | ccccc | ccccc | ccccc | ccccc | ccccc}
\toprule
\multirow{2}{*}{Scene} & \multicolumn{5}{c|}{COLMAP\cite{schonberger2016sfm}} & \multicolumn{5}{c|}{COLMAP\cite{schonberger2016sfm}+DKM\cite{edstedt2023dkm}} & \multicolumn{5}{c|}{COLMAP\cite{schonberger2016sfm}+RoMa\cite{edstedt2024roma}} & \multicolumn{5}{c|}{MAST3R-SfM\cite{duisterhof2025mast3rsfm}} & \multicolumn{5}{c|}{VGGT\cite{vggt2024}} & \multicolumn{5}{c}{\ours{}} \\
\cmidrule(lr){2-6} \cmidrule(lr){7-11} \cmidrule(lr){12-16} \cmidrule(lr){17-21} \cmidrule(lr){22-26} \cmidrule(l){27-31}
& RRA & RTA & ACC & COMP & \rev{AUC} & RRA & RTA & ACC & COMP & \rev{AUC} & RRA & RTA & ACC & COMP & \rev{AUC} & RRA & RTA & ACC & COMP & \rev{AUC} & RRA & RTA & ACC & COMP & \rev{AUC} & RRA & RTA & ACC & COMP & \rev{AUC} \\
& @5$^\circ$ & @5$^\circ$ & @5cm & @5cm & \rev{@5$^\circ$} & @5$^\circ$ & @5$^\circ$ & @5cm & @5cm & \rev{@5$^\circ$} & @5$^\circ$ & @5$^\circ$ & @5cm & @5cm & \rev{@5$^\circ$} & @5$^\circ$ & @5$^\circ$ & @5cm & @5cm & \rev{@5$^\circ$} & @5$^\circ$ & @5$^\circ$ & @5cm & @5cm & \rev{@5$^\circ$} & @5$^\circ$ & @5$^\circ$ & @5cm & @5cm & \rev{@5$^\circ$} \\
\midrule
courtyard\,\rev{(38)} & \hcbr{100} & \hcbr{99.9} & \hc{84.9} & \hc{7.4} & \hcr{89.7} & \hcr{48.9} & \hcr{18.6} & \hc{0} & \hc{0} & \hcr{7.3} & \hcr{20.2} & \hcr{10.7} & \hc{2.3} & \hc{0.4} & \hcr{3.0} & \rev{\hc{62.9}} & \rev{\hc{37.7}} & \rev{\hc{7.2}} & \rev{\hc{6.5}} & \rev{\hc{24.3}} & \hc{94.7} & \hc{89.5} & \hcb{92.6} & \hc{5.1} & \rev{\hc{41.2}} & \rev{\hcb{100.0}} & \rev{\hcb{99.9}} & \hc{74.1} & \hcb{29.1} & \rev{\hcb{89.8}} \\
delivery\_area\,\rev{(44)} & \hcbr{100} & \hcr{99.6} & \hc{64.3} & \hc{5.7} & \hcr{89.0} & \hcr{29.2} & \hcr{25.2} & \hc{4.5} & \hc{0.2} & \hcr{16.5} & \hcr{26.7} & \hcr{21.8} & \hc{3.3} & \hc{0.2} & \hcr{12.2} & \rev{\hc{33.1}} & \rev{\hc{44.0}} & \rev{\hc{4.1}} & \rev{\hc{0.8}} & \rev{\hc{23.3}} & \hc{65.9} & \hc{68.2} & \hcb{90.3} & \hc{0.3} & \rev{\hc{24.0}} & \rev{\hcb{100.0}} & \rev{\hcb{99.8}} & \hc{79.7} & \hcb{34.2} & \rev{\hcb{90.5}} \\
electro\,\rev{(45)} & \hcr{82.8} & \hcr{81.1} & \hc{10.8} & \hc{0.5} & \hcr{71.6} & \hcr{82.8} & \hcr{73.7} & \hc{19.4} & \hc{6.1} & \hcr{50.3} & \hcr{73.3} & \hcr{63.1} & \hc{6.3} & \hc{0.9} & \hcr{37.7} & \rev{\hc{95.6}} & \rev{\hc{87.4}} & \rev{\hc{60.9}} & \rev{\hcb{77.2}} & \rev{\hcb{73.7}} & \hcb{100} & \hcb{93.3} & \hc{50.9} & \hc{13.9} & \rev{\hc{28.1}} & \rev{\hc{95.7}} & \rev{\hc{86.5}} & \hcb{66.0} & \hc{55.2} & \rev{\hc{73.1}} \\
facade\,\rev{(76)} & \hcbr{100} & \hcbr{99.9} & \hc{5.3} & \hc{0.3} & \hcbr{93.2} & \hcr{66.4} & \hcr{49.6} & \hc{3.5} & \hc{0.2} & \hcr{20.2} & \hcr{68.3} & \hcr{58.5} & \hc{3.3} & \hc{0.3} & \hcr{37.6} & \rev{\hc{73.7}} & \rev{\hc{71.1}} & \rev{\hc{5.5}} & \rev{\hc{2.3}} & \rev{\hc{59.1}} & \hc{98.7} & \hc{90.8} & \hc{55.7} & \hc{6.1} & \rev{\hc{62.1}} & \rev{\hcb{100.0}} & \rev{\hc{99.5}} & \hcb{59.8} & \hcb{19.3} & \rev{\hc{91.9}} \\
kicker\,\rev{(31)} & \hcr{93.5} & \hcr{92.9} & \hc{9.3} & \hc{0.3} & \hcr{82.4} & \hcr{87.3} & \hcr{74.2} & \hc{12.0} & \hc{2.0} & \hcr{43.4} & \hcr{49.9} & \hcr{55.5} & \hc{13.8} & \hc{1.4} & \hcr{24.5} & \rev{\hcb{100.0}} & \rev{\hc{99.8}} & \rev{\hc{81.3}} & \rev{\hcb{94.5}} & \rev{\hcb{95.1}} & \hcb{100} & \hcb{100} & \hc{88.1} & \hc{64.5} & \rev{\hc{66.4}} & \rev{\hcb{100.0}} & \rev{\hc{99.1}} & \hcb{94.3} & \hc{47.1} & \rev{\hc{90.3}} \\
meadow\,\rev{(15)} & \hcr{1} & \hcr{1} & \hc{23.6} & \hc{0.1} & \hcr{0} & \hcr{48.6} & \hcr{7.6} & \hc{0} & \hc{0} & \hcr{2.8} & \hcr{57.1} & \hcr{13.3} & \hc{11.0} & \hc{0.2} & \hcr{4.9} & \rev{\hc{53.3}} & \rev{\hc{55.2}} & \rev{\hc{3.3}} & \rev{\hc{3.1}} & \rev{\hcb{44.5}} & \hcb{100} & \hcb{86.7} & \hcb{50.5} & \hcb{52.0} & \rev{\hc{34.8}} & \rev{\hc{86.7}} & \rev{\hc{49.5}} & \hc{19.1} & \hc{3.7} & \rev{\hc{35.9}} \\
office\,\rev{(26)} & \hcbr{100} & \hcr{86.5} & \hc{81.4} & \hc{0.5} & \hcr{63.6} & \hcr{77.8} & \hcr{50.8} & \hc{54.6} & \hc{1.6} & \hcr{34.3} & \hcr{48.3} & \hcr{35.1} & \hc{66.0} & \hc{1.2} & \hcr{22.1} & \rev{\hcb{100.0}} & \rev{\hc{93.8}} & \rev{\hc{72.3}} & \rev{\hcb{85.3}} & \rev{\hcb{74.6}} & \hcb{100} & \hcb{96.2} & \hcb{84.2} & \hc{12.8} & \rev{\hc{37.3}} & \rev{\hc{77.5}} & \rev{\hc{54.8}} & \hc{11.0} & \hc{1.6} & \rev{\hc{43.7}} \\
pipes\,\rev{(14)} & \hcbr{100} & \hcbr{100} & \hc{67.2} & \hc{2.0} & \hcbr{92.5} & \hcbr{100} & \hcr{74.7} & \hc{64.9} & \hc{11.6} & \hcr{49.8} & \hcbr{100} & \hcr{74.7} & \hc{62.6} & \hc{11.4} & \hcr{45.8} & \rev{\hcb{100.0}} & \rev{\hcb{100.0}} & \rev{\hc{79.2}} & \rev{\hcb{81.4}} & \rev{\hc{83.4}} & \hcb{100} & \hc{92.9} & \hcb{98.4} & \hc{5.2} & \rev{\hc{46.9}} & \rev{\hcb{100.0}} & \rev{\hcb{100.0}} & \hc{93.3} & \hc{39.4} & \rev{\hc{90.8}} \\
playground\,\rev{(38)} & \hcbr{100} & \hcbr{99.9} & \hc{40.6} & \hc{8.5} & \hcbr{93.8} & \hcr{89.6} & \hcr{81.7} & \hc{38.2} & \hc{8.0} & \hcr{56.9} & \hcbr{100} & \hcr{92.5} & \hc{41.2} & \hc{23.1} & \hcr{69.9} & \rev{\hc{99.3}} & \rev{\hc{85.5}} & \rev{\hc{9.9}} & \rev{\hc{34.0}} & \rev{\hc{59.1}} & \hc{97.4} & \hc{23.7} & \hc{55.6} & \hc{8.5} & \rev{\hc{10.3}} & \rev{\hcb{100.0}} & \rev{\hc{99.4}} & \hcb{63.1} & \hcb{63.9} & \rev{\hc{89.5}} \\
relief\,\rev{(31)} & \hcbr{100} & \hcr{99.4} & \hcb{96.3} & \hc{18.1} & \hcr{88.5} & \hcr{75.5} & \hcr{57.8} & \hc{63.2} & \hc{26.4} & \hcr{32.4} & \hcr{99.8} & \hcr{81.5} & \hc{68.6} & \hc{32.0} & \hcr{58.8} & \rev{\hc{67.7}} & \rev{\hc{64.1}} & \rev{\hc{33.0}} & \rev{\hc{37.1}} & \rev{\hc{34.8}} & \hcb{100} & \hc{90.3} & \hc{40.6} & \hc{18.9} & \rev{\hc{31.7}} & \rev{\hcb{100.0}} & \rev{\hcb{99.8}} & \hc{93.6} & \hcb{75.9} & \rev{\hcb{91.4}} \\
relief\_2\,\rev{(31)} & \hcbr{100} & \hcbr{100} & \hcb{93.9} & \hc{16.2} & \hcbr{90.3} & \hcbr{100} & \hcr{94.2} & \hc{39.4} & \hc{12.1} & \hcr{63.8} & \hcbr{100} & \hcr{97.8} & \hc{0} & \hc{0} & \hcr{79.4} & \rev{\hcb{100.0}} & \rev{\hc{99.6}} & \rev{\hc{63.7}} & \rev{\hcb{77.1}} & \rev{\hc{82.4}} & \hc{93.6} & \hc{77.4} & \hc{40.5} & \hc{44.4} & \rev{\hc{16.6}} & \rev{\hcb{100.0}} & \rev{\hc{84.1}} & \hc{89.6} & \hc{64.6} & \rev{\hc{54.2}} \\
terrace\,\rev{(23)} & \hcbr{100} & \hcbr{100} & \hc{60.8} & \hc{5.5} & \hcbr{93.0} & \hcbr{100} & \hcr{99.2} & \hc{47.8} & \hc{16.4} & \hcr{82.7} & \hcbr{100} & \hcr{98.8} & \hc{45.4} & \hc{16.1} & \hcr{80.1} & \rev{\hcb{100.0}} & \rev{\hcb{100.0}} & \rev{\hc{47.7}} & \rev{\hcb{71.2}} & \rev{\hc{82.5}} & \hcb{100} & \hcb{100} & \hcb{88.7} & \hc{7.1} & \rev{\hc{33.8}} & \rev{\hcb{100.0}} & \rev{\hcb{100.0}} & \hc{72.4} & \hc{68.7} & \rev{\hc{91.6}} \\
terrains\,\rev{(42)} & \hcbr{100} & \hcbr{99.5} & \hcb{94.0} & \hc{10.2} & \hcbr{91.4} & \hcr{96.4} & \hcr{64.2} & \hc{0} & \hc{0} & \hcr{28.8} & \hcr{99.5} & \hcr{78.4} & \hc{92.8} & \hc{0.1} & \hcr{48.4} & \rev{\hc{40.8}} & \rev{\hc{27.4}} & \rev{\hc{45.5}} & \rev{\hc{2.3}} & \rev{\hc{19.5}} & \hcb{100} & \hc{45.2} & \hc{54.7} & \hc{45.9} & \rev{\hc{14.3}} & \rev{\hcb{100.0}} & \rev{\hc{99.2}} & \hc{92.5} & \hcb{69.0} & \rev{\hc{89.4}} \\
\midrule
\textbf{Mean} & \hcr{90.6} & \hcr{89.2} & \hc{56.3} & \hc{5.8} & \hcbr{79.9} & \hcr{77.1} & \hcr{59.3} & \hc{26.7} & \hc{6.5} & \hcr{37.6} & \hcr{72.6} & \hcr{60.1} & \hc{32.1} & \hc{6.7} & \hcr{40.4} & \rev{\hc{78.9}} & \rev{\hc{74.3}} & \rev{\hc{39.5}} & \rev{\hcb{44.1}} & \rev{\hc{58.2}} & \hc{96.2} & \hc{81.1} & \hc{68.5} & \hc{21.9} & \rev{\hc{34.4}} & \rev{\hcb{96.9}} & \rev{\hcb{90.1}} & \hcb{69.9} & \hc{44.0} & \rev{\hc{78.6}} \\
\bottomrule
\end{tabular}
\end{adjustbox}
\vspace{-0.4cm}
\end{table*}

We evaluate \ours{}'s ability to recover camera poses, uncalibrated intrinsics, and scene geometry from a strict cold start, across diverse environments from object-centric captures to unbounded outdoor scenes. Throughout, we report Relative Rotation and Translation Accuracy (RRA/RTA) at several angular thresholds, \rev{together with the pose AUC, which scores each pose by the larger of its two angular errors. All metrics are defined in \Cref{sec:metrics}.}

\medskip\noindent\textbf{Benchmark Datasets.} \revlabel{b10}We evaluate on four standard benchmarks, treating all image collections as unordered:
\begin{itemize}[noitemsep,topsep=0pt,leftmargin=*]
    \item \textbf{DTU}~\cite{Jensen_2014_CVPR}: Following the protocol from~\cite{truong2022sparf}, we use 15 of the 128 object-centric scenes (49 images each, 735 in total), covering varying levels of difficulty.
    \item \textbf{Mip-NeRF 360}~\cite{barron2022mipnerf360}: 7 real-world indoor and outdoor scenes captured from full 360-degree viewpoints (125--311 images each, 1{,}626 in total).
    \item \textbf{ETH3D}~\cite{schops2017multi}: 13 high-resolution indoor/outdoor scenes (14--76 images each) within complex structures, featuring ground-truth poses alongside reference LiDAR scans for evaluating map accuracy and completeness.
    \item \textbf{IMC2021}~\cite{jin2021image}: 9 in-the-wild, crowdsourced scenes. From each scene's 100-image set we evaluate the standard 175 bundles (100 of 5, 50 of 10, and 25 of 25 views), totalling 1{,}575 multi-view bags, to test robustness against severe occlusions and diverse, uncalibrated intrinsics.
\end{itemize}
\medskip\noindent\textbf{Comparative Baselines.} 
We compare \ours{} against both classical pipelines and deep-learning foundation models. Classical baselines include the standard \textbf{COLMAP}~\cite{schonberger2016sfm} (using default SIFT features) and its dense variants, \textbf{COLMAP + DKM}~\cite{edstedt2023dkm} and \textbf{COLMAP + RoMa}~\cite{edstedt2024roma}. As naively replacing SIFT with dense correspondences often leads to reconstruction failures, we construct robust feature tracks for these variants following the detector-free Structure-from-Motion pipeline~\cite{he2024dsfm}. We further compare against the initialization-free relaxations \textbf{pOSE}~\cite{Hong_2018_CVPR} and \textbf{PoVar}~\cite{weber2024power}. For models utilizing learned geometric priors, we benchmark against \textbf{MASt3R-SfM}~\cite{duisterhof2025mast3rsfm} and \textbf{VGGT}~\cite{vggt2024}. \rev[a6]{We additionally report \textbf{GLOMAP}~\cite{pan2024glomap} and more COLMAP/GLOMAP matcher variants in \Cref{sec:more_baselines}.}

\medskip\noindent\textbf{Implementation Details.} 
To ensure computational tractability, we downsample the dense pairwise matches to a $64 \times 64$ spatial grid (detailed in Appendix~\ref{sec:dense_matching}) and optimize the 3D Gaussian depth and variance fields as $32 \times 32$ parametric maps. \rev{The formulation is matcher-agnostic. \Cref{sec:generalization_dense_matchers} repeats the ETH3D evaluation with RoMa~\cite{edstedt2024roma} and MASt3R~\cite{revaud2024mast3r} front-ends in place of DKM.} Our custom CUDA implementation runs on a single NVIDIA H100 GPU using AdamW (learning rate $0.0016$ for all parameter groups---relative poses, intrinsics, depths, and variances) for 30,000 iterations. \Cref{tab:hyperparams} lists the full set of hyperparameters and schedules. For our cold-start initialization, tree-edge relative poses are set to the identity matrix, field of view to $45^\circ$, and principal points to the image center. \rev{We use a single focal length ($f_x\!=\!f_y$) and keep the principal point fixed at the image center. Only the focal length is optimized.} Initial 3D Gaussian depths $d$ and variances $\sigma^2$ are uniformly set to $1.0$.

\vspace{-0.2cm}
\subsection{Quantitative Results}
\label{sec:results}
\vspace{-0.2cm}

\medskip\noindent\textbf{Performance on ETH3D.} 
\revlabel{b4}As shown in \Cref{tab:aggregated_results_all_horizontal} \rev{(per-scene results and point-cloud metrics in \Cref{tab:per_scene_metrics_compact})}, \ours{} achieves an average Relative Translation Accuracy (RTA@5$^\circ$) of \rev{90.1\%}, outperforming recent foundation models like VGGT and remaining on par with standard COLMAP. While COLMAP performs well with sparse features \rev[a9colmap]{(89.2\%)}, forcing it to process dense correspondences (COLMAP+DKM) results in severe track-building failures, reducing its accuracy to \rev[a9dkm]{59.3\%}. \ours{} circumvents this bottleneck via pairwise Gaussian overlaps. The initialization-free relaxations pOSE and PoVar, which share our cold-start goal, instead collapse on these unconstrained scenes (near-zero accuracy on Mip-NeRF~360 and ETH3D), and dense correspondences degrade rather than aid them. Beyond pose estimation, aligning our reconstructed point clouds to ground-truth LiDAR scans, \ours{} achieves the highest Accuracy (ACC@5cm) of 69.9\% among all methods and a Completeness (COMP@5cm) of 44.0\%, far exceeding classical baselines while remaining competitive with deep models.

\medskip\noindent\textbf{In-the-Wild Scenes with Diverse Intrinsics (IMC2021).} 
\rev{IMC2021 is the benchmark where classical SfM keeps a clear margin: COLMAP reaches RRA@5$^\circ$ 85.3\% and RTA@5$^\circ$ 65.3\% against our 71.2\% and 39.5\%. Moreover, our rotations stay far closer to COLMAP than our translations do. This asymmetry is the signature of the scale--focal length ambiguity (\Cref{fig:focal_depth_am}), which happens on nearly planar scenes. Without intrinsic priors and without a parallax-gated initialization, a cold-start global optimization absorbs residual focal error into camera positions while leaving orientations well constrained.}

\begin{figure}[t]
    \centering
    \captionof{table}{\textbf{Component Ablation.} RTA/RRA under Soft Spatial Constraint (SSC), relative pose optimization (RP), Adaptive Edge Weighting (AEW), and Mirror-Symmetry Regularization (MSR).}
    \label{tab:ablation}
    \vspace{2pt}
    \resizebox{0.8\linewidth}{!}{
        \setlength{\tabcolsep}{4pt}
        \begin{tabular}{cccc | cc | cc}
            \toprule
             & & & & \multicolumn{2}{c|}{ETH3D} & \multicolumn{2}{c}{DTU} \\
            \cmidrule(lr){5-6} \cmidrule(l){7-8}
            SSC & RP & AEW & MSR & RRA@5$^\circ$ & RTA@5$^\circ$ & RRA@5$^\circ$ & RTA@5$^\circ$ \\
            \midrule
            \multicolumn{4}{c|}{Classical BA} & 3.6 & 1.6 & 18.8 & 8.3 \\
            \midrule
            - & - & - & - & 49.9 & 45.1 & 15.1 & 17.3 \\
            - & \checkmark & - & - & 84.0 & 77.3 & 26.7 & 27.2 \\
            \checkmark & - & - & - & 61.9 & 53.9 & 21.7 & 20.9 \\
            \checkmark & \checkmark & - & -& 92.4  & 86.4 & 20.0 & 18.3 \\
            \checkmark & \checkmark & \checkmark & -& \textbf{96.9} & \textbf{90.1}  & 21.4 & 18.4 \\
            \checkmark & \checkmark & \checkmark & \checkmark & \textbf{96.9} & \textbf{90.1} & \textbf{97.9} & \textbf{83.1} \\
            \bottomrule
        \end{tabular}
    }

    \vspace{6pt}
    \includegraphics[width=\linewidth]{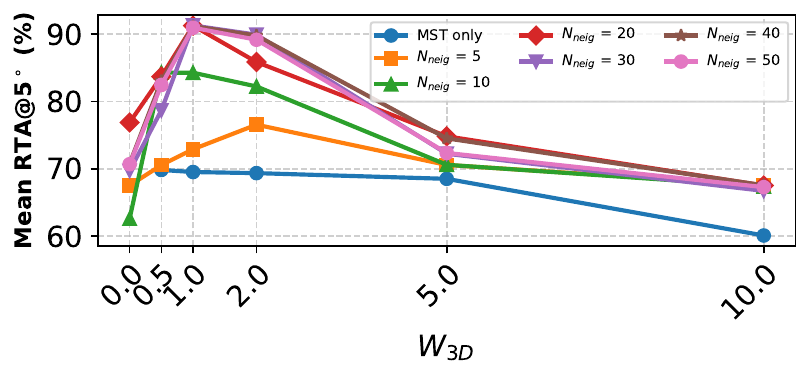}
    \vspace{-0.7cm}
    \captionof{figure}{\textbf{Hyperparameter sensitivity.} Mean ETH3D RTA@5$^\circ$ vs.\ the $\mathcal{L}_{3D}$ weight $W_{3D}$, one curve per view-graph density $N_{\text{neig}}$.}
    \label{fig:ablation_w3d}
    \vspace{-0.4cm}
\end{figure}

\begin{figure}[htb]
    \centering
    \noindent\makebox[\linewidth][l]{\fontsize{2.8}{3.4}\selectfont%
      \hspace*{0.0127\linewidth}%
      \makebox[0.1406\linewidth]{Input images}%
      \makebox[0.1406\linewidth]{COLMAP~\cite{schonberger2016sfm}}%
      \makebox[0.1406\linewidth]{COLMAP~\cite{schonberger2016sfm}+DKM~\cite{edstedt2023dkm}}%
      \makebox[0.1406\linewidth]{COLMAP~\cite{schonberger2016sfm}+RoMa~\cite{edstedt2024roma}}%
      \makebox[0.1406\linewidth]{MASt3R-SfM~\cite{duisterhof2025mast3rsfm}}%
      \makebox[0.1406\linewidth]{VGGT~\cite{vggt2024}}%
      \makebox[0.1406\linewidth]{\ours{}}}\\[1pt]
    \includegraphics[width=\linewidth]{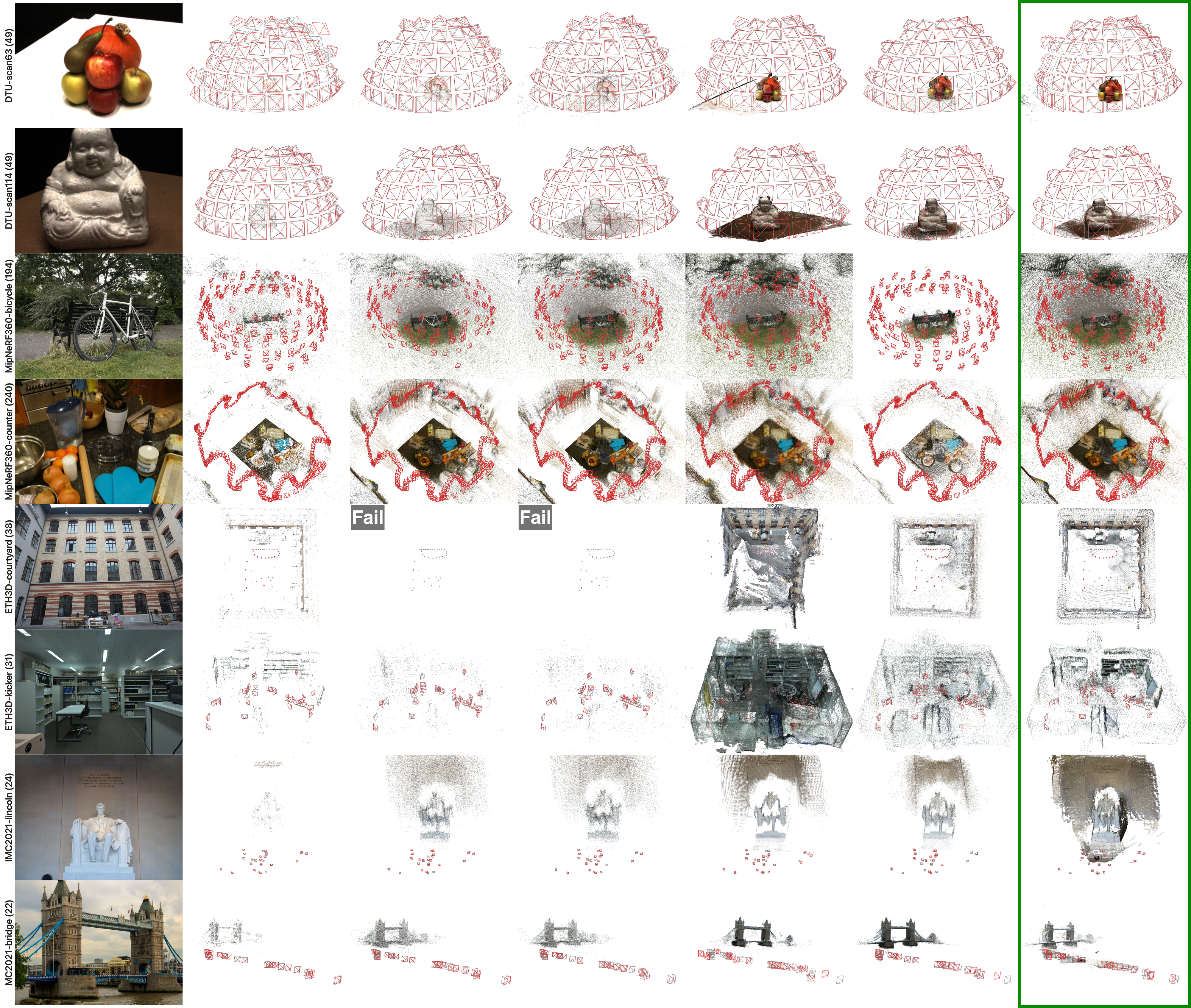}
    \caption{\textbf{Qualitative Reconstructions.} Estimated (red) and ground-truth (grey) camera poses and point clouds across the tested datasets (DTU, Mip-NeRF 360, ETH3D, and IMC2021). Left to right: Input, COLMAP, COLMAP+DKM, COLMAP+RoMa, MASt3R-SfM, VGGT, and \ours{}. While classical methods yield shattered geometries when forced to use dense matches, \ours{} consistently recovers coherent topology and accurate trajectories on par with foundation models.
    }
    \label{fig:qualitative}
\end{figure}

\medskip\noindent\textbf{Performance on DTU and Mip-NeRF 360.} 
Our method scales well to 360-degree and object-centric scenes (\Cref{fig:qualitative}). \ours{} reaches average RRA@5$^\circ$/RTA@5$^\circ$ of \rev{99.6/98.5\%} on Mip-NeRF~360 and \rev{97.9/83.1\%} on DTU---the latter despite the initialization challenges of object-centric framing and textureless backgrounds. These figures are on par with COLMAP and its dense variants (COLMAP+DKM, COLMAP+RoMa) on these largely saturated benchmarks, even though \ours{} runs from a strict cold start.

\subsection{Ablation Studies}
\label{sec:ablation}

\medskip\noindent\textbf{Optimization Components.} 
We ablate our core components on ETH3D and DTU (\Cref{tab:ablation}). With the 2D reprojection NLL alone, the Gaussian landmarks already raise ETH3D RTA@5$^\circ$ from 1.6\% (classical BA) to 45.1\%. Adding the Soft Spatial Constraint (SSC) lifts this to 53.9\%. \revlabel{b2}Relative-pose (RP) optimization adds about 32 points with SSC (86.4\%) as well as without it (77.3\%). Adaptive Edge Weighting (AEW) then reaches \rev{90.1\%} by filtering bad edges. On DTU, Mirror-Symmetry Regularization (MSR) is the critical catalyst: center-concentrated features over textureless backgrounds yield ambiguous early gradients prone to inverted (mirror) solutions, and by maintaining dual hypotheses and selecting the true geometry via the NLL loss, MSR lifts DTU RTA@5$^\circ$ from 18.4\% to \rev{83.1\%}.

\medskip\noindent\textbf{View-Graph Density and Weight of $\mathcal{L}_{3D}$.} 
We further ablate view-graph density---the number of neighbor edges per node ($N_{\text{neig}}$)---and the soft-constraint weight $W_{3D}$ (\Cref{fig:ablation_w3d}). An overly sparse graph (MST only or $N_{\text{neig}}{=}5$) under-constrains the optimization and yields the lowest accuracy, while density helps up to $N_{\text{neig}}{=}20$ and then plateaus. Both removing the 3D term ($W_{3D}{=}0$) and overweighting it ($W_{3D}{\geq}5$) lower accuracy. We thus default to $N_{\text{neig}}{=}20$ and $W_{3D}{=}1.0$ (balancing accuracy and runtime). \rev[b12]{ Results are also insensitive to tree topology (\Cref{sec:mst_robustness}).}

\begin{figure}[t]
    \centering
    \includegraphics[width=0.49\linewidth]{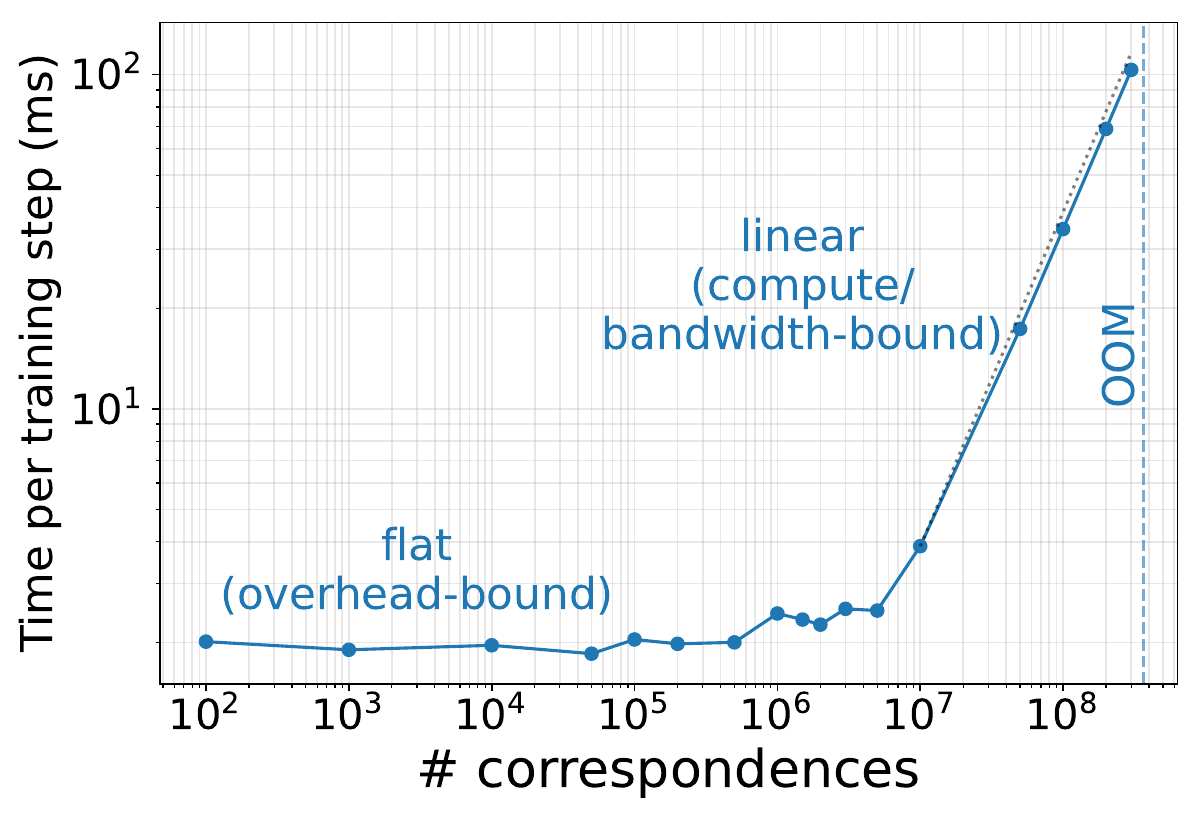}
    \hfill
    \includegraphics[width=0.49\linewidth]{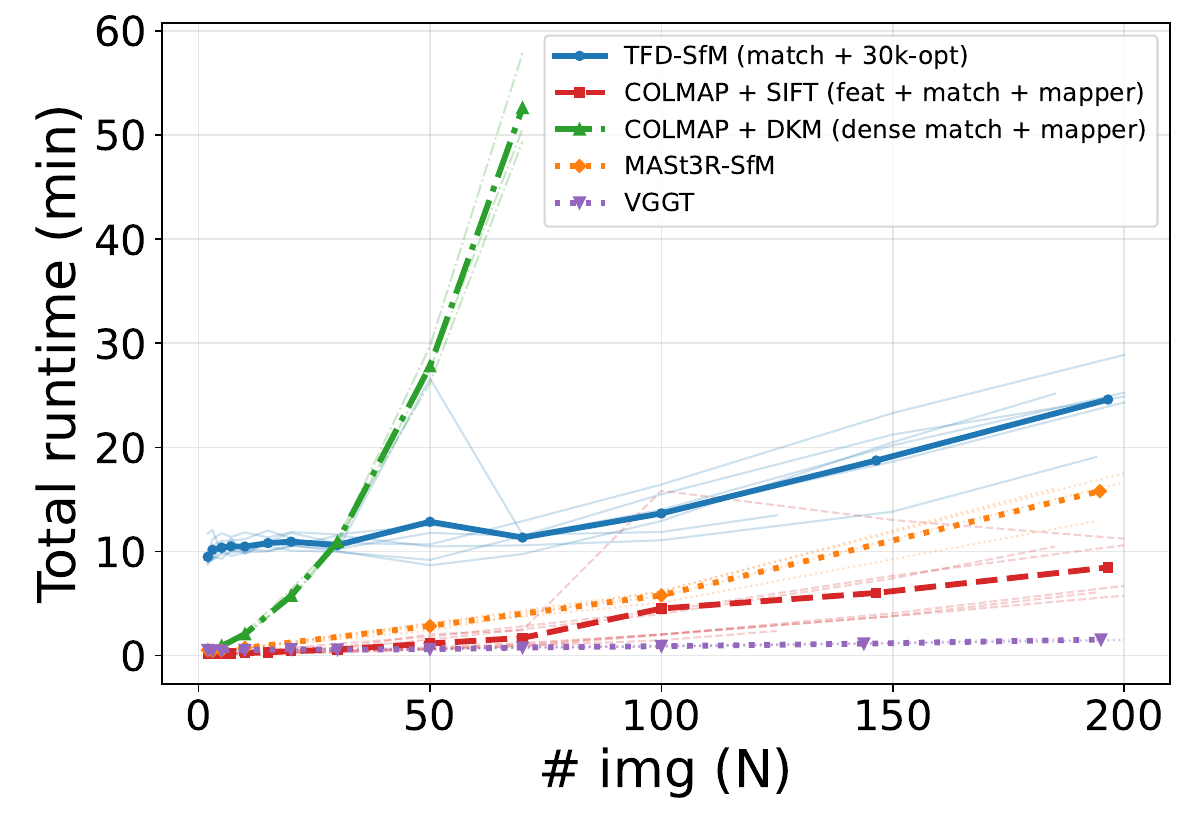}
    \vspace{-0.2cm}
    \caption{\textbf{Runtime analysis.} (Left) Time per optimization step vs.\ correspondence count $M$ (log--log axes). The dotted line marks linear scaling in $M$, and the dashed line marks the out-of-memory limit of a 95\,GB GPU. (Right) End-to-end runtime vs.\ number of input views $N$ on Mip-NeRF~360 for \ours{} (blue), COLMAP (red), COLMAP+DKM (green), MASt3R-SfM (orange), and VGGT (purple). Transparent lines show single scenes and solid lines the mean.}
    \label{fig:runtime}
\end{figure}

\medskip\noindent\textbf{Computational Efficiency.}
\ours{} scales favorably in both the correspondence count $M$ and the view count $N$. At the correspondence counts of real scenes, one optimization step takes ${\sim}2$\,ms, a cost set by fixed per-step overhead rather than by $M$. The Mip-NeRF~360 scenes ($125$--$311$ images) yield only ${\sim}0.8$--$6$\,M correspondences, and the step cost grows linearly with $M$ only well beyond this range (measured up to ${\sim}10^8$ on a 95\,GB GPU, \Cref{fig:runtime}, left). A full run, from building the matches to the end of the 30k optimization steps, takes ${\sim}10$--$25$\,min on Mip-NeRF~360 (\Cref{fig:runtime}, right). COLMAP+DKM, the directly comparable dense pipeline, registers images one at a time and runs bundle adjustment over the growing model after each registration. Its runtime grows superlinearly (${\sim}O(N^2)$), from ${\sim}1$\,min at $N{=}5$ to ${\sim}50$\,min at $N{=}70$, while \ours{} needs only ${\sim}10$\,min at $N{=}70$. Sparse COLMAP is cheaper, but it recovers only a sparse point cloud, not the dense geometry \ours{} targets. \rev{Among the deep-scene-prior baselines, feed-forward VGGT is the cheapest method overall and stays nearly flat in $N$, while MASt3R-SfM grows steadily.}
\section{Conclusion}
\label{sec:conclusion}
We introduced \ours{}, which reformulates Bundle Adjustment from rigid point tracking into an initialization-free optimization by relaxing tracks into 3D Gaussians under a unified Negative Log-Likelihood objective, with adaptive edge weighting and a dual-hypothesis mechanism for the mirror-symmetry ambiguity. Together these broaden the narrow basins of attraction of classical solvers and assimilate noisy dense matches through their learned variance rather than discarding them, registering views on which dense track-building collapses and recovering geometry far more complete than classical pipelines, at a predictable computational cost. \rev[b5]{Its remaining limitations are the focal-depth ambiguity on near-planar scenes, a pinhole camera model without lens distortion, and a long first-order convergence tail, each of which opens a concrete direction for future work.} We hope \ours{} gives a practical foundation for initialization-free 3D reconstruction in unstructured environments.

\paragraph{Acknowledgments.}
\rev{This work was supported by the ERC Advanced Grant SIMULACRON and the DFG project CR~250/26-1 ``4D-YouTube: 4D-Rekonstruktion aus Gelegenheitsvideos mit tiefen raum-zeitlichen Repr\"asentationen''.}

{
    \small
    \bibliographystyle{ieeenat_fullname}

}

\appendix
\clearpage
\raggedbottom  %
\section{Appendix}\label{appendix}

\subsection{Spanning Tree and View Graph Construction}
\label{sec:viewgraph}

\begin{figure*}[t]
\hrule height.8pt depth0pt \kern2pt
\captionsetup{type=algorithm,name=Algorithm}
\caption{\rev{Spanning Tree and View Graph Construction}}
\label{alg:mst}
\kern2pt\hrule height.4pt depth0pt \kern2pt
    \begin{algorithmic}[1]
    \Require Set of uncalibrated images $\mathcal{I} = \{I_1, I_2, \dots, I_N\}$
    \Ensure \rev{View Graph $\mathcal{G}=\{V,E\}$, Spanning Tree $\mathcal{T}=\{V,E_{tree}\}$,} Similarity Matrix $\mathbf{S}$, Parent Mapping $P$
    
        \Procedure{BuildST}{$\mathcal{I}$}
            \State $V \gets \{1, 2, \dots, N\}$, $E_{tree} \gets \emptyset$
            \State $\mathbf{S}_{ij} = |\Call{LoFTR}{I_i, I_j}| \quad \forall i, j \in V$ \Comment{Number of correspondences}
            \State $\mathbf{S} \gets \mathbf{S} / \max(\mathbf{S})$ \Comment{Normalize similarities to $[0,1]$}
            \State $r \gets \arg\max_{i \in V} \sum_{j} \mathbf{S}_{ij}$ \Comment{Designate most connected node as root $r$}
            \State $V_{proc} \gets \{r\}$, $V_{unproc} \gets V \setminus \{r\}$
            
            \While{$V_{unproc} \neq \emptyset$}
                \State $(u, v) \gets \arg\max_{i \in V_{unproc}, j \in V_{proc}} \mathbf{S}_{ij}$ \Comment{Highest single-edge similarity}
                \State $P(u) \gets v$ \Comment{Record $v$ as parent of $u$}
                \State $E_{tree} \gets E_{tree} \cup \{(u, v)\}$
                \State $V_{proc} \gets V_{proc} \cup \{u\}$, $V_{unproc} \gets V_{unproc} \setminus \{u\}$
            \EndWhile
            \State \rev{$\mathcal{T} \gets \{V, E_{tree}\}$} \Comment{\rev{The spanning tree, our kinematic chain for pose parametrization}}
            \Statex
            \State \rev{$E \gets E_{tree}$} \Comment{\rev{Augment the tree into the view graph}}
            \ForAll{\rev{$i \in V$}}
                \State \rev{$E \gets E \cup \{(i,j) : j \in \textsc{TopK}(\mathbf{S}_{i\cdot},\, N_{\text{neig}}-1)\}$}
            \EndFor
            \State \rev{$\mathcal{G} \gets \{V, E\}$} \Comment{\rev{The view graph over which the NLL cost is optimized}}
            
            \State \Return \rev{$\mathcal{G}, \mathcal{T},$} $\mathbf{S}, P$
        \EndProcedure
    \end{algorithmic}
\kern2pt\hrule height.8pt depth0pt
\end{figure*}

\ours{} operates on a view graph $\mathcal{G} = \{V, E\}$ and its spanning tree $\mathcal{T}$ (Figure~\ref{fig:overview}). Each node $i \in V$ is a camera with absolute pose $\mathbf{T}_{ri}$ and unknown intrinsics $\mathbf{K}_i$. \rev{Algorithm~\ref{alg:mst} builds both graphs in three steps.}

\rev{First, a coarse low-resolution LoFTR~\cite{sun2021loftr} pass scores all $O(N^2)$ image pairs. It keeps a single scalar per pair, the number of confident matches, which forms the normalized similarity matrix $\mathbf{S}$.}

\rev{Second, we grow the spanning tree greedily on $\mathbf{S}$. The most connected image becomes the root. Each step attaches the unprocessed node with the highest similarity to any processed node and records its parent in $P$. The tree edges $E_{tree}$ define the kinematic chain of \Cref{sec:kinematic}.}

\rev{Third, we augment the tree into the view graph. Every node adds edges to its $N_{\text{neig}}-1$ most similar neighbours under $\mathbf{S}$ (we use $N_{\text{neig}}=20$).}

\subsection{Spanning-Tree Robustness}
\label{sec:mst_robustness}
\begin{figure}[H]
    \centering
    \includegraphics[width=\linewidth]{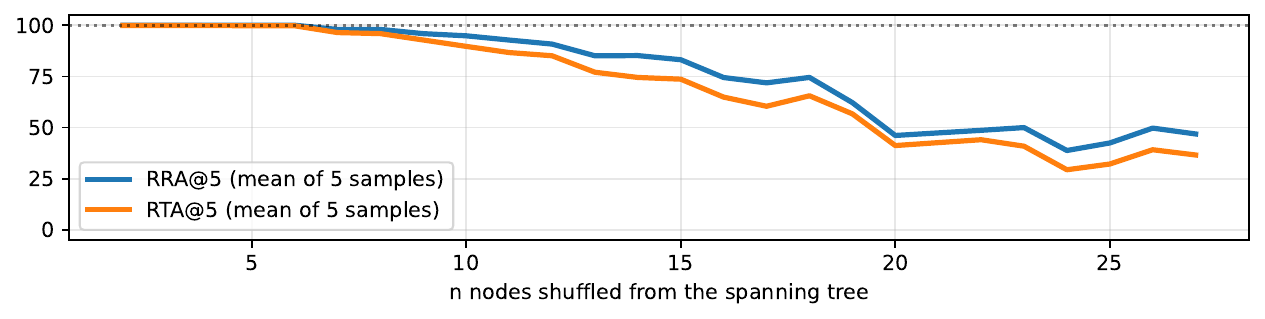}
    \caption{\textbf{Spanning-tree robustness.} RRA/RTA@5$^\circ$ versus the number of tree nodes whose parent is randomly reassigned (mean of $5$ seeds).}
    \label{fig:mst_robustness}
\end{figure}
To probe how much the reconstruction depends on the specific spanning tree, on the ETH3D \emph{courtyard} scene we start from the deterministic tree and randomly reassign the parent of $n$ non-root nodes, then re-run the full optimization, sweeping $n$ and averaging over $5$ random seeds (\Cref{fig:mst_robustness}). RRA/RTA@5$^\circ$ remain at $100\%$ for $n\le 7$ and degrade only gracefully beyond. The perturbation is deliberately adversarial---it reroutes high-overlap image pairs through unrelated cameras, a regime the similarity-driven tree construction avoids in practice---so the default operating point sits well inside the robust plateau. The choice of root node has no measurable effect.

\subsection{Optimization Details}
\label{sec:optim_details}
All free parameters---per-camera relative poses, focal lengths, and the per-view depth and variance fields---are optimized jointly with AdamW from the cold start over $30{,}000$ iterations. \Cref{tab:hyperparams} consolidates the hyperparameters and schedules. Convergence behaviour (a rapid phase followed by a long first-order tail) is analyzed in \Cref{fig:convergence_plots}.

\begin{table}[t]
\centering
\footnotesize
\caption{\textbf{Optimization hyperparameters and schedules.}}
\label{tab:hyperparams}
\begin{tabular}{@{}p{0.40\linewidth} p{0.52\linewidth}@{}}
\toprule
Setting & Value \\
\midrule
Optimizer & AdamW ($\epsilon{=}10^{-15}$, weight decay $0$) \\
Iterations & $30{,}000$ \\
Learning rate (all groups) & $0.0016$ \\
Betas (poses, depth, variance) & $(0.9,\,0.999)$ \\
Betas (focal) & $(0.5,\,0.999)$ \\
Depth \& variance maps & $32{\times}32$ per view, bilinear \\
Adaptive edge weighting & frozen for first $5{,}000$ iters, then EMA $\alpha{=}10^{-4}$ \\
Mirror reg.\ $\lambda$ & annealed $100\!\to\!0$ over first $2{,}000$ iters \\
Initialization & identity poses; $\mathrm{fov}=45^\circ$; centered principal point; $d{=}1$; $\sigma^2{=}1$ \\
\bottomrule
\end{tabular}
\end{table}

\subsection{Adaptive Edge Weighting: Schedule and Initialization}
\label{sec:impl}

\noindent\textbf{Local $z$-score and reliability.} The edge reliability of \Cref{sec:adaptedge} is built from a local $z$-score: we standardize each edge NLL against the edges incident to node $i$ (its view-graph neighbors $\mathcal{N}(i) = \{k : e_{ik} \in E\}$). With per-node mean $\mu_i = \frac{1}{|\mathcal{N}(i)|}\sum_{k \in \mathcal{N}(i)} \mathcal{L}^{ik}$ and standard deviation $\nu_i = \big[\frac{1}{|\mathcal{N}(i)|}\sum_{k \in \mathcal{N}(i)} (\mathcal{L}^{ik} - \mu_i)^2\big]^{1/2}$ (distinct from the Gaussian variance $\sigma_u^2$ of \Cref{sec:objective}), the standardized edge loss is $\bar{\mathcal{L}}_{ij} = (\mathcal{L}^{ij} - \mu_i)/\nu_i$, negative for better-than-average edges. A logistic map averaged over the edge's two endpoints (each standardized at its own node) gives the per-edge reliability:
\begin{equation}
    w_{ij} = \tfrac{1}{2}\operatorname{sigmoid}(-\bar{\mathcal{L}}_{ij}) + \tfrac{1}{2}\operatorname{sigmoid}(-\bar{\mathcal{L}}_{ji}),
\end{equation}
so $w_{ij}\to1$ for reliable edges and $\to0$ for outliers.

\medskip\noindent\textbf{Temporal smoothing.} The reliability scores are temporally smoothed into a reliability $W_{ij}^{(t)}$ via an exponential moving average (EMA) with update rate $\alpha^{(t)}$:
\begin{equation}
\begin{aligned}
    W_{ij}^{(t)} &= (1 - \alpha^{(t)})\,W_{ij}^{(t-1)} + \alpha^{(t)}\, w_{ij}, \\
    \alpha^{(t)} &= \begin{cases} 0, & t \le 5000 \\ 10^{-4}, & t > 5000 \end{cases}
\end{aligned}
\end{equation}
The delayed schedule ($\alpha^{(t)} = 0$ for the first $5000$ iterations) freezes the weights while the geometry stabilizes, preventing premature pruning of edges. The weights are initialized from the visual similarity $\mathbf{S}_{ij}$ (Algorithm~\ref{alg:mst}), with auxiliary (non-tree) edges down-weighted:
\begin{equation}
    W_{ij}^{(0)} = \mathbf{S}_{ij} \left(\tfrac{1}{N_{\text{neig}}}\right)^{\mathbb{I}_{\text{aux}}(i,j)},
\end{equation}
where the indicator $\mathbb{I}_{\text{aux}}(i,j)$ equals $1$ if $e_{ij}$ is an auxiliary neighborhood edge and $0$ if it is a primary tree branch.

\subsection{Dense Correspondence Generation and Filtering}
\label{sec:dense_matching}

Given the established view graph $\mathcal{G} = \{V, E\}$ (from Section~\ref{sec:viewgraph}), we generate dense correspondences for each valid edge $e_{ij} \in E$ by computing bidirectional dense matching fields via \textbf{DKM}~\cite{edstedt2023dkm}. To ensure computational tractability and to guarantee a spatially uniform distribution of geometric constraints—which prevents highly textured local regions from dominating the optimization landscape—we uniformly sample these dense fields using a $64 \times 64$ spatial grid. 

For each sampled coordinate $\mathbf{c}_i$ in image $i$, we obtain its putative match in image $j$ via the forward flow $F_{i \to j}$ such that $\mathbf{c}_j = F_{i \to j}(\mathbf{c}_i)$. To filter occlusions and severe matching outliers, the candidate pair $(\mathbf{c}_i, \mathbf{c}_j)$ is added to the pairwise correspondence set $\mathcal{C}_{ij}$ only if it satisfies the forward-backward consistency constraint:
\begin{equation}
    \|F_{j \to i}(\mathbf{c}_j) - \mathbf{c}_i\|_2 \leq \tau
\end{equation}
where $F_{j \to i}$ denotes the backward flow mapping from image $j$ back to $i$, and the error threshold is set to $\tau = 5$ pixels. The global correspondence set for bundle adjustment is the union of all filtered matches: $\mathcal{C} = \bigcup_{e_{ij} \in E} \mathcal{C}_{ij}$.

\subsection{Toy-Model Construction}
\label{sec:toy_setup}
The controlled instance of \Cref{fig:toy} is fully synthetic. We place $100$ ground-truth landmarks on the surface of a sphere (radius $1.3$) via a Fibonacci lattice for even coverage, centred at the scene centroid. The $N$ cameras sit on a frontal orbit (radius $4$) at a fixed $15^\circ$ angular spacing, so the viewpoint arc spans $(N{-}1)\times 15^\circ$ and widens toward the full $360^\circ$ ring as cameras are added one at a time. Each landmark is co-visible in a contiguous run of $4$--$6$ cameras (resampled so every consecutive camera pair shares at least one landmark), and the view graph contains every camera pair with at least one shared landmark. Correspondences are formed by projecting the ground-truth landmarks through the ground-truth camera poses with a pinhole model and adding i.i.d.\ isotropic Gaussian pixel noise of standard deviation $\sigma$ (one fixed realization per scene). The associations are therefore exact and contain no gross outliers. We sweep $\sigma\in\{0,1,4,8\}$\,px and average over $5$ random seeds (independent visibility and noise draws). Crucially, the ground-truth poses are used \emph{only} to synthesize these observations: every run optimizes \Cref{eq:unified} from a strict cold start (camera poses at the identity, unit depth, no metric priors), so the recovered geometry reflects the basin of attraction of the objective alone. The $2{\times}2$ ablation crosses pose parametrization (independent per-camera poses vs.\ a chained relative-pose composition) with the landmark model (one shared 3D point per track, i.e.\ classical BA, vs.\ our per-observation 3D Gaussians).

\subsection{Evaluation Metrics}
\label{sec:metrics}

To evaluate the estimated camera trajectory and the reconstructed 3D map, we employ standard accuracy and completeness metrics. Let $\mathbb{I}(\cdot)$ denote the indicator function, which evaluates to $1$ if the condition is true and $0$ otherwise.

\vspace{1mm}\noindent\textbf{Camera Pose Metrics.} Given $N$ evaluated poses, let $R_{gt}^{(i)}, R_{est}^{(i)} \in SO(3)$ and $t_{gt}^{(i)}, t_{est}^{(i)} \in \mathbb{R}^3$ be the ground truth and estimated rotation matrices and translation vectors for the $i$-th pose. Their rotation and translation angular errors are
{\small
\begin{equation}
    e_R^{(i)} = \arccos\left( \frac{\text{tr}({R_{gt}^{(i)}}^T R_{est}^{(i)}) - 1}{2} \right)
\end{equation}
\begin{equation}
    e_t^{(i)} = \arccos\left( \frac{{t_{gt}^{(i)}}^T t_{est}^{(i)}}{\lVert t_{gt}^{(i)} \rVert_2 \lVert t_{est}^{(i)} \rVert_2} \right)
\end{equation}
}
We measure the \textit{Relative Rotation Accuracy} (RRA) and \textit{Relative Translation Accuracy} (RTA) as the fraction of poses whose angular error falls strictly below a predefined tolerance threshold $x^\circ$ (e.g., $5^\circ$ or $10^\circ$), which dictates the strictness of the evaluation:
{\small
\begin{equation}
    \text{RRA}@x^\circ = \frac{1}{N} \sum_{i=1}^{N} \mathbb{I}\left( e_R^{(i)} < x \right), \quad
    \text{RTA}@x^\circ = \frac{1}{N} \sum_{i=1}^{N} \mathbb{I}\left( e_t^{(i)} < x \right)
\end{equation}
}

\begin{revblock}
\vspace{1mm}\noindent\textbf{Pose AUC.} \revlabel{b7}RRA and RTA each report accuracy at a single
tolerance and score rotation and translation separately. To summarise both over a range of
tolerances, we additionally report the pose AUC. The joint error of a pose is
\begin{equation}
    e^{(i)} = \max\!\left(e_R^{(i)},\; e_t^{(i)}\right).
\end{equation}
With the accuracy curve $\mathrm{Acc}(\tau) = \frac{1}{N}\sum_{i=1}^{N}\mathbb{I}\!\left(e^{(i)} < \tau\right)$,
the AUC at threshold $x^\circ$ is its integral, normalised so that a perfect estimate scores $1$:
\begin{equation}
    \text{AUC}@x^\circ = \frac{1}{x}\int_{0}^{x} \mathrm{Acc}(\tau)\, \mathrm{d}\tau ,
\end{equation}
computed by trapezoidal integration over the sorted errors and reported as a percentage.
\end{revblock}

\vspace{1mm}\noindent\textbf{3D Reconstruction Metrics.} Let $\mathcal{P}_{est}$ and $\mathcal{P}_{gt}$ be the estimated and ground truth 3D point clouds. Prior to evaluation, we first align the two maps using robust Iterative Closest Point (ICP)\cite{besl1992icp}. We evaluate \textit{Accuracy} (ACC@$x$ cm) as the fraction of estimated points within $x$ cm of the true geometry, and \textit{Completeness} (COMP@$x$ cm) as the fraction of true points within $x$ cm of the estimated geometry:
{\small
\begin{equation}
    \text{ACC}@x\text{ cm} = \frac{1}{|\mathcal{P}_{est}|} \sum_{p \in \mathcal{P}_{est}} \mathbb{I}\left(\min_{q \in \mathcal{P}_{gt}} \lVert p - q \rVert_2 < x\right)
\end{equation}
\begin{equation}
    \text{COMP}@x\text{ cm} = \frac{1}{|\mathcal{P}_{gt}|} \sum_{q \in \mathcal{P}_{gt}} \mathbb{I}\left(\min_{p \in \mathcal{P}_{est}} \lVert p - q \rVert_2 < x\right)
\end{equation}
}

\subsection{Visualization of the Evolution of 3D Gaussians and poses}
\label{sec:vis_evolution}

\begin{figure}[t]
    \centering
    \includegraphics[width=\linewidth]{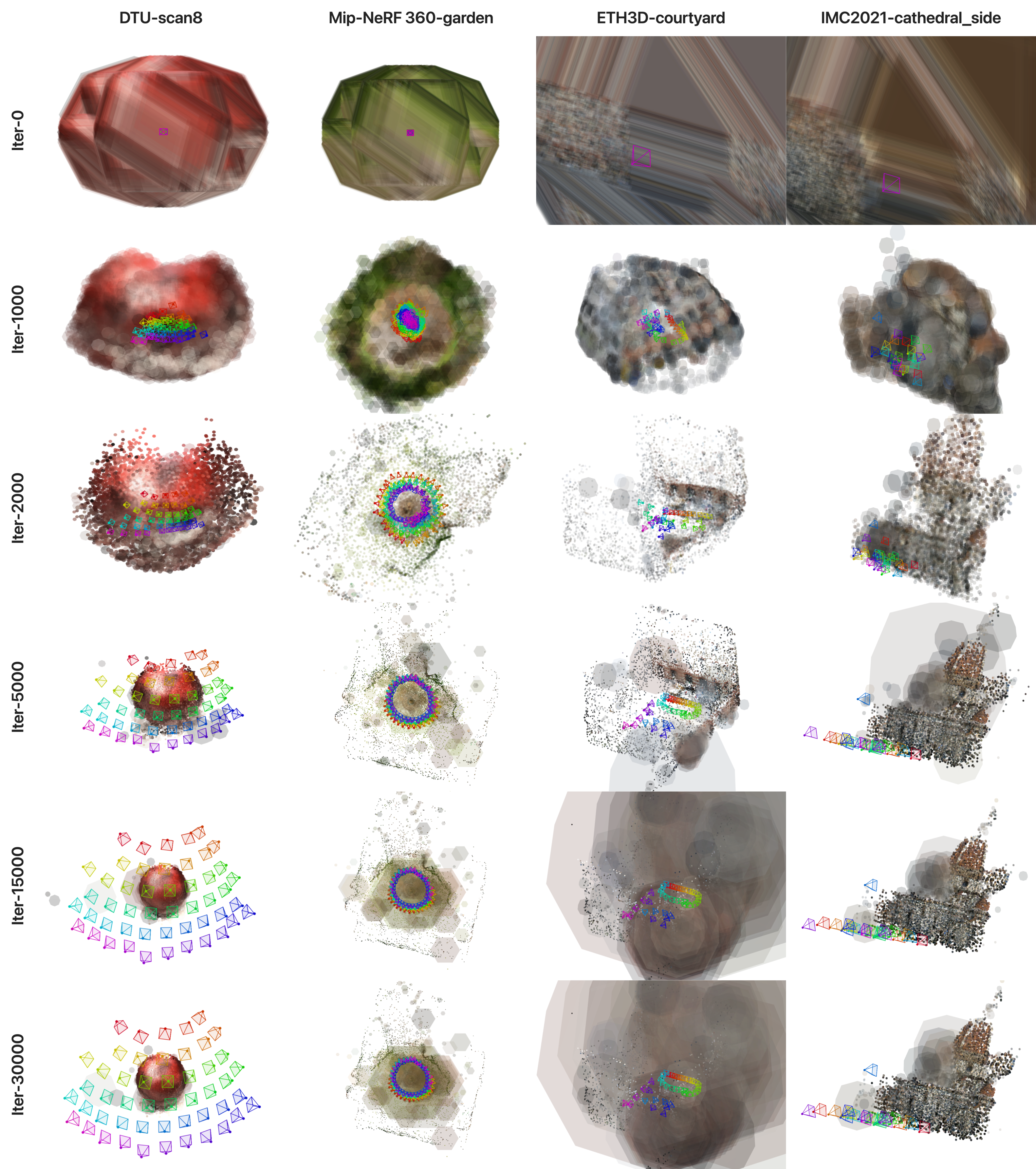}
    \caption{\textbf{Joint evolution of camera poses and 3D Gaussians.} Initialized from a cold start, the camera poses, intrinsics, and 3D Gaussians are jointly optimized. Shown from left to right are representative scenes from the four evaluated datasets: DTU, Mip-NeRF 360, ETH3D, and IMC2021. \textit{Note: Due to rendering constraints, visualizations display a randomly subsampled set of 10,000 Gaussians per scene, depicting isotropic Gaussians as polygonal spheres rather than perfectly smooth spheres.}}
    \label{fig:evo_poses_gaussians}
    \vspace{-0.2cm}
\end{figure}

We visualize the joint optimization process of our framework across four distinct datasets (DTU, Mip-NeRF 360, ETH3D, and IMC2021) in Figure~\ref{fig:evo_poses_gaussians}. The system initializes all cameras at an identity pose, with 3D Gaussians positioned one unit away along the optical axis and assigned an initial variance of one unit (Iter-0). As the optimization steps progress, the cameras fan out to discover their true global poses while the Gaussians concurrently coalesce to form the underlying scene geometry. The joint optimization typically reaches stable convergence by iteration 15,000, with minimal visual changes observed by iteration 30,000.

Crucially, this visualization highlights the robust outlier-handling mechanism inherent to our approach. In essence, these high-variance Gaussians act as probabilistic shock absorbers. Rather than forcing incorrect correspondences into a rigid, physically invalid geometry---which catastrophically shatters classical track-building---the framework accommodates them by expanding their spatial variance. This phenomenon is visually evident in the later optimization stages (e.g., Iter-15,000 onwards in the ETH3D and IMC2021 scenes), where gross outliers manifest as large, highly diffuse spheres. By naturally increasing their variance to down-weight their influence, the framework visually demonstrates our core claim: volumetric uncertainty effectively smooths the highly non-convex loss landscape, preventing catastrophic local minima and enabling stable, prior-free convergence.

\subsection{Dual-Hypothesis Convergence Dynamics}
\label{sec:dual_hypothesis_dynamics}

\noindent\textbf{Mirror-regularization schedule.} The weight $\lambda_{\text{mir}}(t) = \lambda_{reg} \max\!\left(1 - \tfrac{t}{t_{\text{mir}}},\, 0\right)$ in \Cref{eq:reg_loss} decays linearly to zero over an early warm-up (values in \Cref{tab:hyperparams}), after which the secondary hypothesis refines without constraint.

\begin{figure}[htb]
    \centering
    \includegraphics[width=\linewidth]{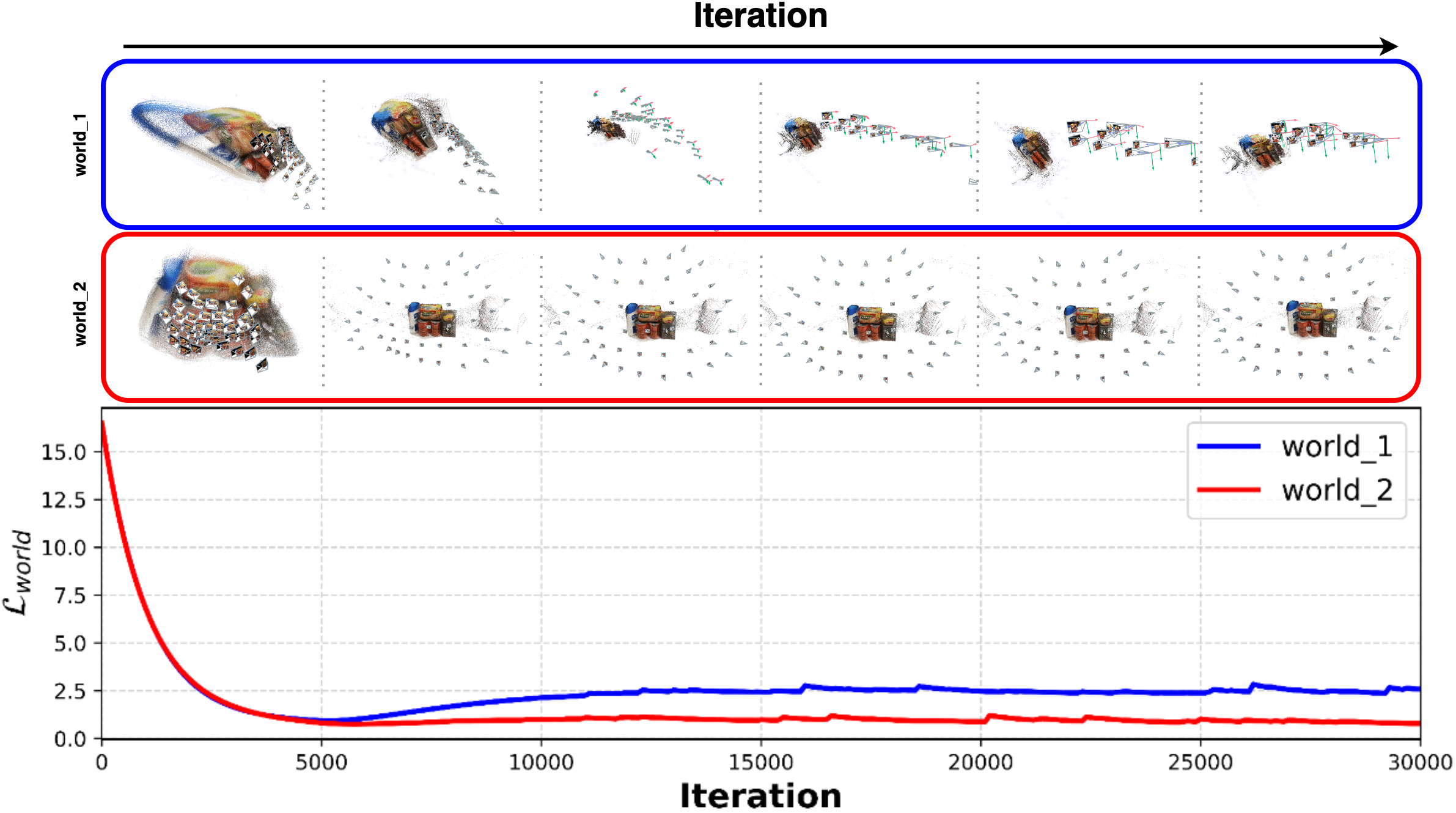}
    \caption{\textbf{Evolution of the two hypothesis worlds and their corresponding loss $\mathcal{L}_{\text{world}}$.} The mirrored \textit{world\_1} (blue) is rejected into a higher-loss minimum; the true \textit{world\_2} (red) descends to convergence. Solid lines: EMA over the raw stochastic data (semi-transparent).}
    \label{fig:dual_world_loss}
    \vspace{-0.2cm}
\end{figure}

To justify the need for optimizing two concurrent scenes, we analyze the evolution of the negative logarithmic likelihood loss ($\mathcal{L}_{\text{world}}$) for both the primary and the mirrored hypotheses. As illustrated in Figure~\ref{fig:dual_world_loss}, at the very beginning of the optimization from a cold start (iterations 0 to $\sim$5,000), the loss function cannot distinguish which world represents the true geometry. Both hypotheses undergo an almost identical, rapid descent. During this early phase, the sparse, noisy correspondences are geometrically ambiguous, meaning both the true structure and its mirror-reflected counterpart appear equally valid to the optimizer. 

However, a clear divergence emerges near iteration 5,000. As the 3D Gaussians densify and multi-view constraints become strictly enforced across wider baselines, the geometric contradictions of the reflected (mirror) geometry become apparent. The loss of the mirrored hypothesis (\textit{world\_1}, blue) increases as it fails to reconcile the conflicting views, ultimately becoming trapped in a higher-loss local minimum characterized by heavily distorted camera poses and corrupted reconstruction. Conversely, the true geometric world (\textit{world\_2}, red) smoothly continues its descent, converging into a correct, consistent topology. 

This delayed disambiguation definitively justifies the dual-world approach and highlights a fundamental advantage of \ours{} over greedy classical pipelines. Whereas traditional SfM prematurely commits to a geometric structure based on brittle initial tracks---often locking irreversibly into a mirrored local minimum---our probabilistic formulation maintains a state of geometric superposition. Crucially, the true geometry (\textit{world\_2}) is preserved via the mirror-symmetry regularization (Eq.~\ref{eq:reg_loss}). If the framework merely committed to a single geometry, the inherent ambiguity of the initial matches could trap the optimization in an unrecoverable mirror-symmetric solution. By optimizing both worlds in parallel, \ours{} safely defers this critical decision until the global topology is sufficiently dense to mathematically reject the spurious mirror solution (as demonstrated by the divergence of \textit{world\_1}).

\subsection{Generalization Across Dense Matchers}
\label{sec:generalization_dense_matchers}

\begin{table}[htb]
\centering
\caption{\textbf{Matcher generalization within \ours{} \rev{on ETH3D, and MASt3R-SfM as a reference}.} Relative Rotation Accuracy (RRA@5$^\circ$), Relative Translation Accuracy (RTA@5$^\circ$), Accuracy at 5cm (ACC@5cm), Completeness at 5cm (COMP@5cm) \rev{and pose AUC@5$^\circ$} for \ours{} driven by three different dense frontends (DKM, RoMa \rev{and MASt3R}), \rev{alongside MASt3R-SfM's own pipeline on the same scenes.}}
\label{tab:diff_dense_matcher}
\begin{adjustbox}{max width=\linewidth}
\setlength{\tabcolsep}{3pt} 
\begin{tabular}{l | ccccc | ccccc | ccccc | ccccc}
\toprule
\multirow{2}{*}{Scene} & \multicolumn{5}{c|}{\rev{MASt3R-SfM\cite{duisterhof2025mast3rsfm}}} & \multicolumn{5}{c|}{\rev{\ours{} (MASt3R)}} & \multicolumn{5}{c|}{\ours{} (DKM)} & \multicolumn{5}{c}{\ours{} (RoMa)} \\
\cmidrule(lr){2-6} \cmidrule(lr){7-11} \cmidrule(lr){12-16} \cmidrule(l){17-21}
& \rev{RRA} & \rev{RTA} & \rev{ACC} & \rev{COMP} & \rev{AUC} & \rev{RRA} & \rev{RTA} & \rev{ACC} & \rev{COMP} & \rev{AUC} & RRA & RTA & ACC & COMP & \rev{AUC} & RRA & RTA & ACC & COMP & \rev{AUC} \\
& \rev{@5$^\circ$} & \rev{@5$^\circ$} & \rev{@5cm} & \rev{@5cm} & \rev{@5$^\circ$} & \rev{@5$^\circ$} & \rev{@5$^\circ$} & \rev{@5cm} & \rev{@5cm} & \rev{@5$^\circ$} & @5$^\circ$ & @5$^\circ$ & @5cm & @5cm & \rev{@5$^\circ$} & @5$^\circ$ & @5$^\circ$ & @5cm & @5cm & \rev{@5$^\circ$} \\
\midrule
courtyard & \rev{\hc{62.9}} & \rev{\hc{37.7}} & \rev{\hc{7.2}} & \rev{\hc{6.5}} & \rev{\hc{24.3}} & \rev{\hcb{100.0}} & \rev{\hc{98.9}} & \rev{\hc{54.7}} & \rev{\hc{19.8}} & \rev{\hc{80.9}} & \rev{\hcb{100.0}} & \rev{\hcb{99.9}} & \hc{74.1} & \hc{29.1} & \rev{\hcb{89.8}} & \rev{\hcb{100.0}} & \rev{\hc{98.3}} & \hcb{81.2} & \hcb{32.3} & \rev{\hc{87.0}} \\
delivery\_area & \rev{\hc{33.1}} & \rev{\hc{44.0}} & \rev{\hc{4.1}} & \rev{\hc{0.8}} & \rev{\hc{23.3}} & \rev{\hc{79.0}} & \rev{\hc{76.2}} & \rev{\hc{10.7}} & \rev{\hc{2.9}} & \rev{\hc{59.8}} & \rev{\hcb{100.0}} & \rev{\hc{99.8}} & \hc{79.7} & \hc{34.2} & \rev{\hc{90.5}} & \rev{\hcb{100.0}} & \rev{\hcb{100.0}} & \hcb{84.9} & \hcb{54.2} & \rev{\hcb{92.6}} \\
electro & \rev{\hc{95.6}} & \rev{\hcb{87.4}} & \rev{\hc{60.9}} & \rev{\hcb{77.2}} & \rev{\hc{73.7}} & \rev{\hc{87.0}} & \rev{\hc{83.6}} & \rev{\hc{42.4}} & \rev{\hc{24.3}} & \rev{\hc{70.5}} & \rev{\hcb{95.7}} & \rev{\hc{86.5}} & \hc{66.0} & \hc{55.2} & \rev{\hc{73.1}} & \rev{\hc{91.2}} & \rev{\hc{86.3}} & \hcb{67.2} & \hc{53.0} & \rev{\hcb{77.4}} \\
facade & \rev{\hc{73.7}} & \rev{\hc{71.1}} & \rev{\hc{5.5}} & \rev{\hc{2.3}} & \rev{\hc{59.1}} & \rev{\hcb{100.0}} & \rev{\hc{98.8}} & \rev{\hc{42.7}} & \rev{\hc{10.9}} & \rev{\hc{83.0}} & \rev{\hcb{100.0}} & \rev{\hcb{99.5}} & \hc{59.8} & \hcb{19.3} & \rev{\hc{91.9}} & \rev{\hcb{100.0}} & \rev{\hc{99.2}} & \hcb{62.4} & \hc{18.8} & \rev{\hcb{92.1}} \\
kicker & \rev{\hcb{100.0}} & \rev{\hcb{99.8}} & \rev{\hc{81.3}} & \rev{\hcb{94.5}} & \rev{\hcb{95.1}} & \rev{\hcb{100.0}} & \rev{\hc{99.3}} & \rev{\hc{89.5}} & \rev{\hc{40.7}} & \rev{\hc{87.0}} & \rev{\hcb{100.0}} & \rev{\hc{99.1}} & \hc{94.3} & \hc{47.1} & \rev{\hc{90.3}} & \rev{\hcb{100.0}} & \rev{\hc{98.9}} & \hcb{95.9} & \hc{45.8} & \rev{\hc{90.2}} \\
meadow & \rev{\hc{53.3}} & \rev{\hc{55.2}} & \rev{\hc{3.3}} & \rev{\hc{3.1}} & \rev{\hc{44.5}} & \rev{\hcb{100.0}} & \rev{\hcb{58.1}} & \rev{\hcb{81.8}} & \rev{\hcb{5.4}} & \rev{\hcb{50.0}} & \rev{\hc{86.7}} & \rev{\hc{49.5}} & \hc{19.1} & \hc{3.7} & \rev{\hc{35.9}} & \rev{\hc{41.0}} & \rev{\hc{15.2}} & \hc{4.0} & \hc{0.0} & \rev{\hc{5.0}} \\
office & \rev{\hcb{100.0}} & \rev{\hcb{93.8}} & \rev{\hc{72.3}} & \rev{\hcb{85.3}} & \rev{\hcb{74.6}} & \rev{\hcb{100.0}} & \rev{\hc{85.8}} & \rev{\hcb{79.8}} & \rev{\hc{44.9}} & \rev{\hc{56.7}} & \rev{\hc{77.5}} & \rev{\hc{54.8}} & \hc{11.0} & \hc{1.6} & \rev{\hc{43.7}} & \rev{\hc{77.5}} & \rev{\hc{55.1}} & \hc{15.0} & \hc{2.4} & \rev{\hc{43.0}} \\
pipes & \rev{\hcb{100.0}} & \rev{\hcb{100.0}} & \rev{\hc{79.2}} & \rev{\hcb{81.4}} & \rev{\hc{83.4}} & \rev{\hc{55.0}} & \rev{\hc{46.1}} & \rev{\hc{88.0}} & \rev{\hc{1.7}} & \rev{\hc{24.7}} & \rev{\hcb{100.0}} & \rev{\hcb{100.0}} & \hc{93.3} & \hc{39.4} & \rev{\hc{90.8}} & \rev{\hcb{100.0}} & \rev{\hcb{100.0}} & \hcb{94.0} & \hc{32.4} & \rev{\hcb{90.9}} \\
playground & \rev{\hc{99.3}} & \rev{\hc{85.5}} & \rev{\hc{9.9}} & \rev{\hc{34.0}} & \rev{\hc{59.1}} & \rev{\hc{25.9}} & \rev{\hc{17.8}} & \rev{\hcb{100.0}} & \rev{\hc{0.0}} & \rev{\hc{1.6}} & \rev{\hcb{100.0}} & \rev{\hc{99.4}} & \hc{63.1} & \hcb{63.9} & \rev{\hc{89.5}} & \rev{\hcb{100.0}} & \rev{\hcb{99.7}} & \hc{12.5} & \hc{1.0} & \rev{\hcb{94.1}} \\
relief & \rev{\hc{67.7}} & \rev{\hc{64.1}} & \rev{\hc{33.0}} & \rev{\hc{37.1}} & \rev{\hc{34.8}} & \rev{\hcb{100.0}} & \rev{\hc{99.1}} & \rev{\hc{91.0}} & \rev{\hc{74.3}} & \rev{\hc{83.8}} & \rev{\hcb{100.0}} & \rev{\hcb{99.8}} & \hc{93.6} & \hc{75.9} & \rev{\hcb{91.4}} & \rev{\hcb{100.0}} & \rev{\hc{99.3}} & \hcb{96.5} & \hcb{77.3} & \rev{\hc{90.6}} \\
relief\_2 & \rev{\hcb{100.0}} & \rev{\hc{99.6}} & \rev{\hc{63.7}} & \rev{\hcb{77.1}} & \rev{\hc{82.4}} & \rev{\hcb{100.0}} & \rev{\hc{99.6}} & \rev{\hc{85.6}} & \rev{\hc{71.9}} & \rev{\hc{81.6}} & \rev{\hcb{100.0}} & \rev{\hc{84.1}} & \hc{89.6} & \hc{64.6} & \rev{\hc{54.2}} & \rev{\hcb{100.0}} & \rev{\hcb{100.0}} & \hcb{94.2} & \hc{75.8} & \rev{\hcb{90.8}} \\
terrace & \rev{\hcb{100.0}} & \rev{\hcb{100.0}} & \rev{\hc{47.7}} & \rev{\hcb{71.2}} & \rev{\hc{82.5}} & \rev{\hcb{100.0}} & \rev{\hcb{100.0}} & \rev{\hc{71.5}} & \rev{\hc{62.3}} & \rev{\hc{83.9}} & \rev{\hcb{100.0}} & \rev{\hcb{100.0}} & \hc{72.4} & \hc{68.7} & \rev{\hc{91.6}} & \rev{\hcb{100.0}} & \rev{\hcb{100.0}} & \hcb{76.4} & \hc{52.9} & \rev{\hcb{93.0}} \\
terrains & \rev{\hc{40.8}} & \rev{\hc{27.4}} & \rev{\hc{45.5}} & \rev{\hc{2.3}} & \rev{\hc{19.5}} & \rev{\hcb{100.0}} & \rev{\hc{94.0}} & \rev{\hc{93.4}} & \rev{\hc{68.7}} & \rev{\hc{67.5}} & \rev{\hcb{100.0}} & \rev{\hcb{99.2}} & \hc{92.5} & \hc{69.0} & \rev{\hcb{89.4}} & \rev{\hcb{100.0}} & \rev{\hc{98.5}} & \hcb{94.0} & \hcb{69.5} & \rev{\hc{87.0}} \\
\midrule
\textbf{Mean} & \rev{\hc{78.9}} & \rev{\hc{74.3}} & \rev{\hc{39.5}} & \rev{\hcb{44.1}} & \rev{\hc{58.2}} & \rev{\hc{88.2}} & \rev{\hc{81.3}} & \rev{\hcb{71.6}} & \rev{\hc{32.9}} & \rev{\hc{63.9}} & \rev{\hcb{96.9}} & \rev{\hcb{90.1}} & \hc{69.9} & \hc{44.0} & \rev{\hc{78.6}} & \rev{\hc{93.1}} & \rev{\hc{88.5}} & \hc{67.6} & \hc{39.6} & \rev{\hcb{79.5}} \\
\bottomrule
\end{tabular}
\end{adjustbox}
\vspace{-0.4cm}
\end{table}

To demonstrate that the robustness of \ours{} stems from its probabilistic formulation rather than a specific network architecture, we substitute our default DKM\cite{edstedt2023dkm} frontend with RoMa\cite{edstedt2024roma}. We evaluate this on the ETH3D high-resolution benchmark, utilizing its precise ground-truth poses and millimeter-accurate LiDAR scans to rigorously assess both pose estimation and 3D reconstruction quality.

As shown in Table~\ref{tab:diff_dense_matcher}, \ours{} maintains highly consistent pose accuracy across most scenes when using RoMa. Although the aggregate mean performance drops compared to DKM, this decrease is heavily skewed by isolated failure cases (e.g., the \textit{meadow} scene) rather than a systemic decline. Therefore, a per-scene evaluation provides a more accurate representation of the framework's adaptability to different noise distributions.

Notably, when both matchers succeed, RoMa generally yields superior 3D map accuracy and completeness. However, because RoMa is approximately five times slower than DKM, we selected DKM for our primary experiments to maintain a practical balance between computational efficiency and geometric fidelity. The soft spatial constraints assimilate dense noise regardless of the underlying frontend, so \ours{}'s robustness is a property of the probabilistic formulation rather than of a specific matcher.

\subsection{Robustness to Extreme Outliers and Noise}
\label{sec:noise_robustness}

\begin{figure}[H]
    \centering
    \includegraphics[width=0.49\linewidth]{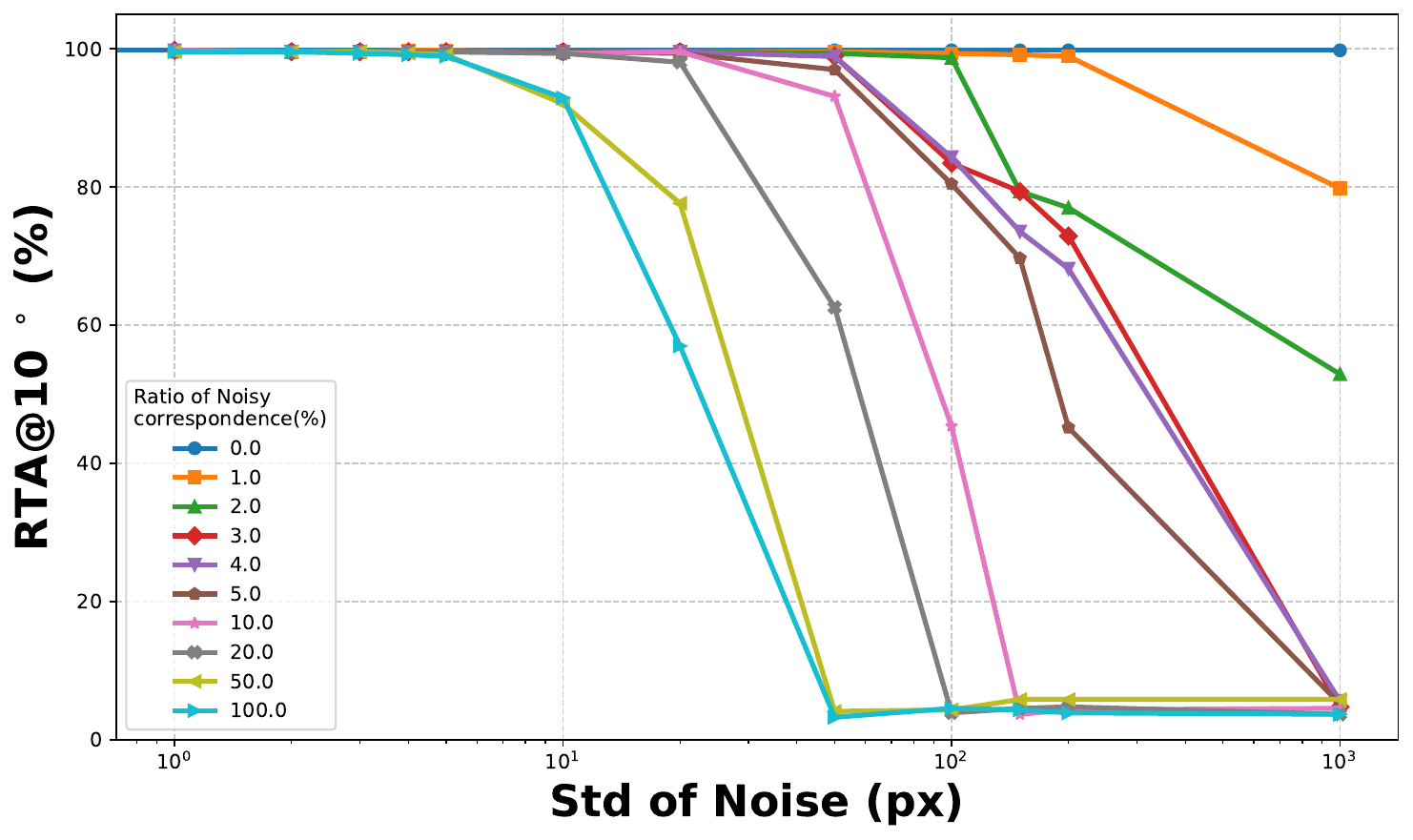}
    \hfill
    \includegraphics[width=0.49\linewidth]{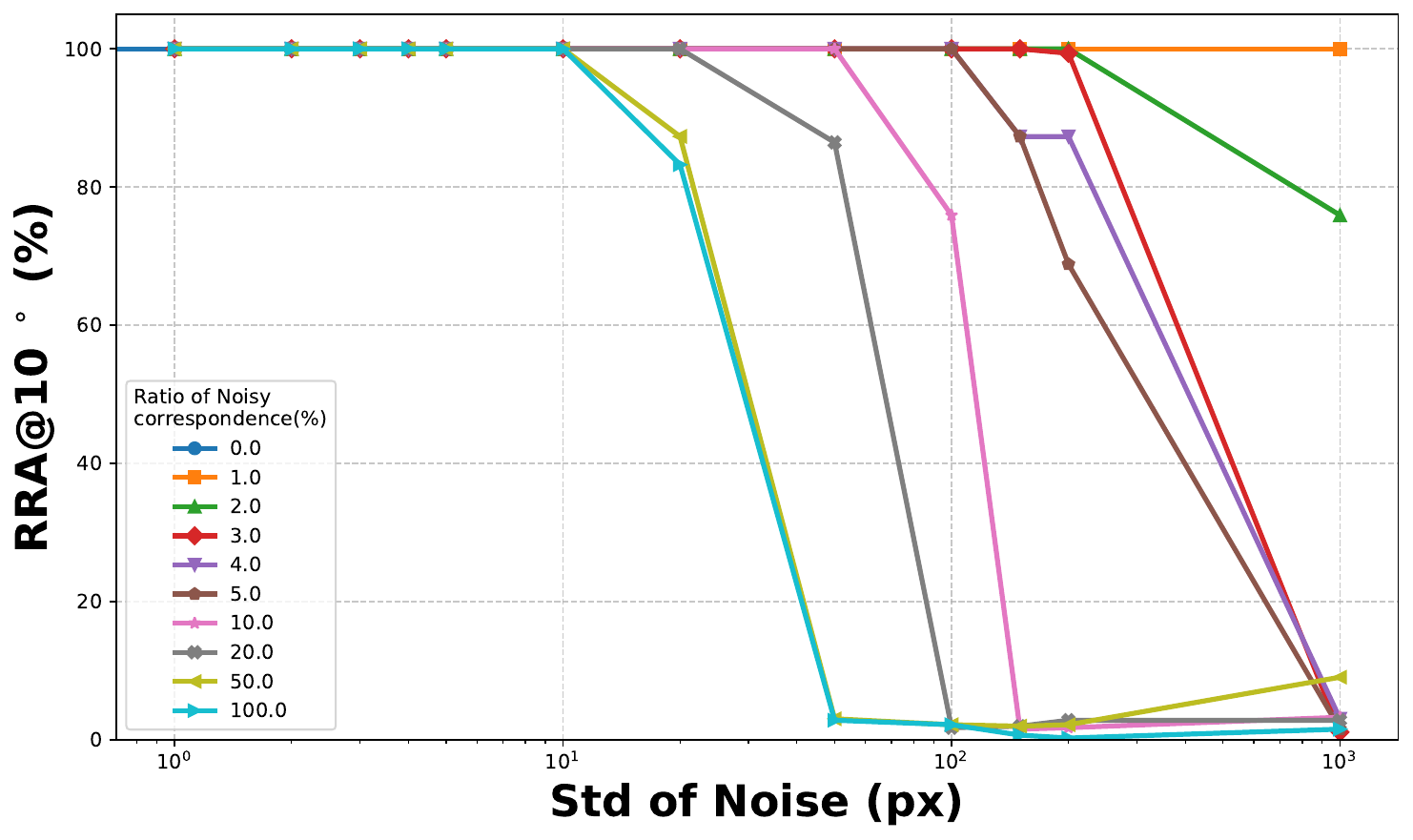}
    \caption{\textbf{Robustness to Controlled Noise.} Relative Translation Accuracy (RTA@$10^\circ$, \textbf{left}) and Relative Rotation Accuracy (RRA@$10^\circ$, \textbf{right}) evaluated across varying noise magnitudes ($\sigma$, x-axis) and ratios ($r$, colored lines) on the ETH3D \textit{kicker} scene. The method maintains high accuracy under both widespread mild noise and sparse gross outliers, degrading only when both parameters are extreme.}
    \label{fig:noise_robustness}
    \vspace{-0.5cm}
\end{figure}

To evaluate robustness to corrupted correspondences, we inject controlled perturbations before optimization. The experiment is governed by two parameters: the \emph{noise ratio} $r\!\in\![0,100]\%$, denoting the fraction of correspondences corrupted per image pair, and the \emph{noise magnitude} $\sigma$ (in pixels), denoting the standard deviation of the applied perturbation.

To each selected target coordinate $\mathbf{u}_{\text{j}}$, we apply noise as follows:
\begin{equation}
  \tilde{\mathbf{u}}_{\text{j}} = \mathbf{u}_{\text{j}} + \boldsymbol{\epsilon}, \qquad \boldsymbol{\epsilon} \sim \mathcal{N}(\mathbf{0}, \sigma^{2}\mathbf{I}_{2}).
\end{equation}

We perform this experiment on the \textit{kicker} scene from the ETH3D dataset. We sweep $r$ from $0\%$ to $100\%$ and $\sigma$ from $0$ to $1000$\,px,  yielding a two-dimensional grid of operating conditions. As illustrated in Figure~\ref{fig:noise_robustness}, this design evaluates both Relative Translation Accuracy (RTA@$10^\circ$) and Relative Rotation Accuracy (RRA@$10^\circ$) across two distinct failure modes: \emph{widespread mild noise} (high~$r$, low~$\sigma$) and \emph{sparse gross outliers} (low~$r$, high~$\sigma$).

The results clearly demonstrate strong tolerance in both regimes, with rotation and translation accuracy exhibiting nearly identical behavior. Under widespread mild noise, even when $100\%$ of the correspondences are corrupted ($r=100\%$), the performance remains completely unaffected up to a standard deviation of $\sigma = 10$\,px. Conversely, the method shows remarkable resilience to sparse gross outliers: when the noise ratio is \rev{$1\%$} or lower, both RTA and RRA remain near $100\%$ even under massive perturbations up to $\sigma = 200$\,px. Naturally, the system eventually breaks down under the most adversarial conditions---when both the percentage of outliers is high ($r \ge 50\%$) and the magnitude of the noise is extreme ($\sigma \ge 50$\,px)---resulting in a sharp decline in overall pose accuracy.

\subsection{Isotropic vs. Anisotropic Gaussians}
\label{sec:iso_aniso}

\begin{table}[htb]
\vspace{-0.4cm}
\centering
\caption{\textbf{Comparison of Isotropic and Anisotropic 3D Gaussians within \ours{}.} Evaluation on the ETH3D dataset for Relative Rotation Accuracy (RRA@5$^\circ$), Relative Translation Accuracy (RTA@5$^\circ$), Accuracy at 5cm (ACC@5cm), and Completeness at 5cm (COMP@5cm). Using the standard anisotropic formulation causes early overfitting to incorrect poses, severely degrading both localization and reconstruction performance.}
\label{tab:iso_vs_aniso}
\begin{adjustbox}{max width=\linewidth}
\setlength{\tabcolsep}{6pt} 
\begin{tabular}{cccc cccc}
\toprule
\multicolumn{4}{c}{Isotropic 3D Gaussians (Ours)} & \multicolumn{4}{c}{Anisotropic 3D Gaussians} \\
\cmidrule(r){1-4} \cmidrule(l){5-8}
RRA & RTA & ACC & COMP & RRA & RTA & ACC & COMP \\
@5$^\circ$ & @5$^\circ$ & @5cm & @5cm & @5$^\circ$ & @5$^\circ$ & @5cm & @5cm \\
\midrule
\rev{\textbf{96.9}} & \rev{\textbf{90.1}} & \textbf{69.9} & \textbf{44.0} & 0.4 & 0.0 & 4.3 & 3.7 \\
\bottomrule
\end{tabular}
\end{adjustbox}
\end{table}

\begin{figure}[htb]
    \centering
    \includegraphics[width=\linewidth]{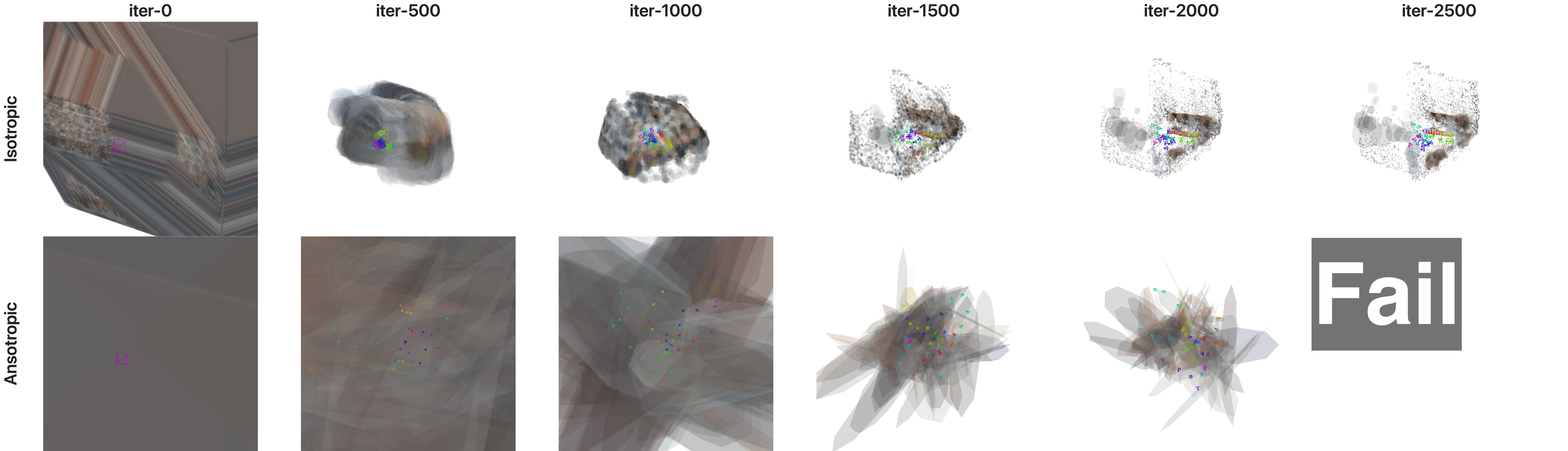}
    \caption{\textbf{Evolution of Isotropic vs. Anisotropic 3D Gaussians during joint optimization.} \textbf{Top:} isotropic Gaussians converge into coherent structure. \textbf{Bottom:} anisotropic Gaussians overfit early poses, stretching into degenerate shapes and failing.}
    \label{fig:evo_aniso}
    \vspace{-0.2cm}
\end{figure}

In \ours{}, we restrict our representation to isotropic 3D Gaussians. As Table~\ref{tab:iso_vs_aniso} demonstrates on the ETH3D dataset, enabling standard anisotropic scaling severely degrades both pose estimation and 3D reconstruction (e.g., RTA@5$^\circ$ drops from \rev{90.1} to 0.0). 

This failure stems from the joint optimization of camera poses and geometry. During early iterations, estimated poses are highly inaccurate. Exploiting their high degrees of freedom, anisotropic Gaussians rapidly overfit to these erroneous viewpoints, stretching into degenerate shapes to minimize immediate photometric loss (Figure~\ref{fig:evo_aniso}, Bottom). As poses update, these distorted, highly directional Gaussians become invalid from new viewpoints, causing severe loss spikes that trap the optimization in local minima. 

Conversely, isotropic Gaussians project symmetrically from all directions (Figure~\ref{fig:evo_aniso}, top). This spherical symmetry prevents premature geometric collapse and provides smoother gradients, stabilizing the joint optimization of poses and geometry.

\subsection{\rev{Limitations and Future Work}}
\label{sec:limitations}
\label{sec:limits_main}

While our approach demonstrates strong robustness to extreme outliers, there are some limitations.

\begin{figure}[tb]
    \centering
    \includegraphics[width=\linewidth]{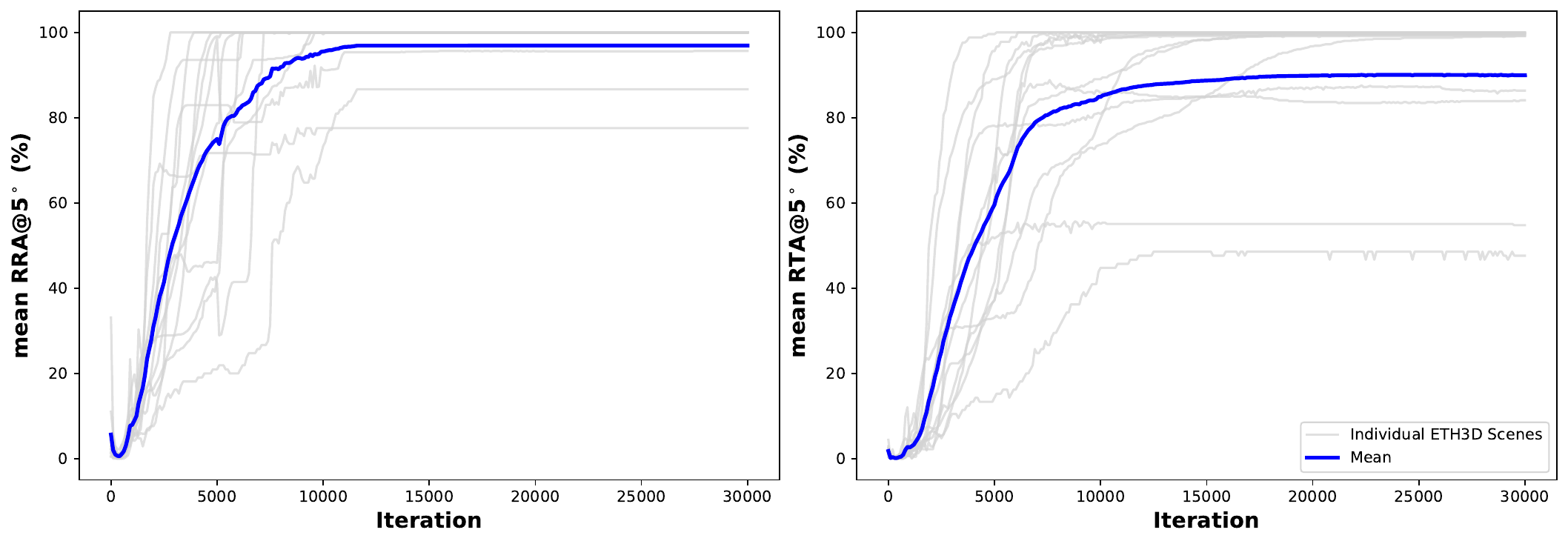}
    \caption{\textbf{Convergence Analysis on ETH3D.} Mean Relative Rotation Accuracy (RRA@5$^\circ$) and Relative Translation Accuracy (RTA@5$^\circ$) across 30,000 iterations. Grey lines represent individual scenes in ETH3D dataset, while the blue line denotes the mean of them.}
    \label{fig:convergence_plots}
    \vspace{-0.2cm}
\end{figure}
\vspace{1mm}\noindent\textbf{First-Order Optimization.} Unlike second-order methods (e.g., Levenberg-Marquardt) that leverage Hessian approximations for rapid convergence near the minimum, our highly scalable first-order solver exhibits a distinct "long tail" during final refinement. However, by accepting this slower asymptotic convergence, \ours{} entirely avoids the prohibitive $\mathcal{O}(N^3)$ memory complexity of Hessian inversions. This deliberate trade-off is what enables our framework to optimize dense, unconstrained view graphs without succumbing to the Out-Of-Memory (OOM) failures that frequently plague classical dense BA. \rev{As shown in Figure~\ref{fig:convergence_plots}, mean RRA@5$^\circ$ saturates within the first $\sim$12,000 iterations, while mean RTA@5$^\circ$ continues to accrue marginal gains along the long tail until roughly iteration 25,000.} Consequently, we establish 30,000 iterations as the standard for all our experiments to guarantee full convergence.

\vspace{1mm}\noindent\rev[b5app]{\textbf{Camera Model.} \ours{} assumes a simple pinhole camera: a single focal length ($f_x\!=\!f_y$) with the principal point fixed at the image centre. Lens distortion is therefore not modelled, and images with pronounced radial distortion---wide-angle or action-camera footage---must be undistorted beforehand.}

\begin{figure}[htb]
    \centering
    \includegraphics[width=\linewidth]{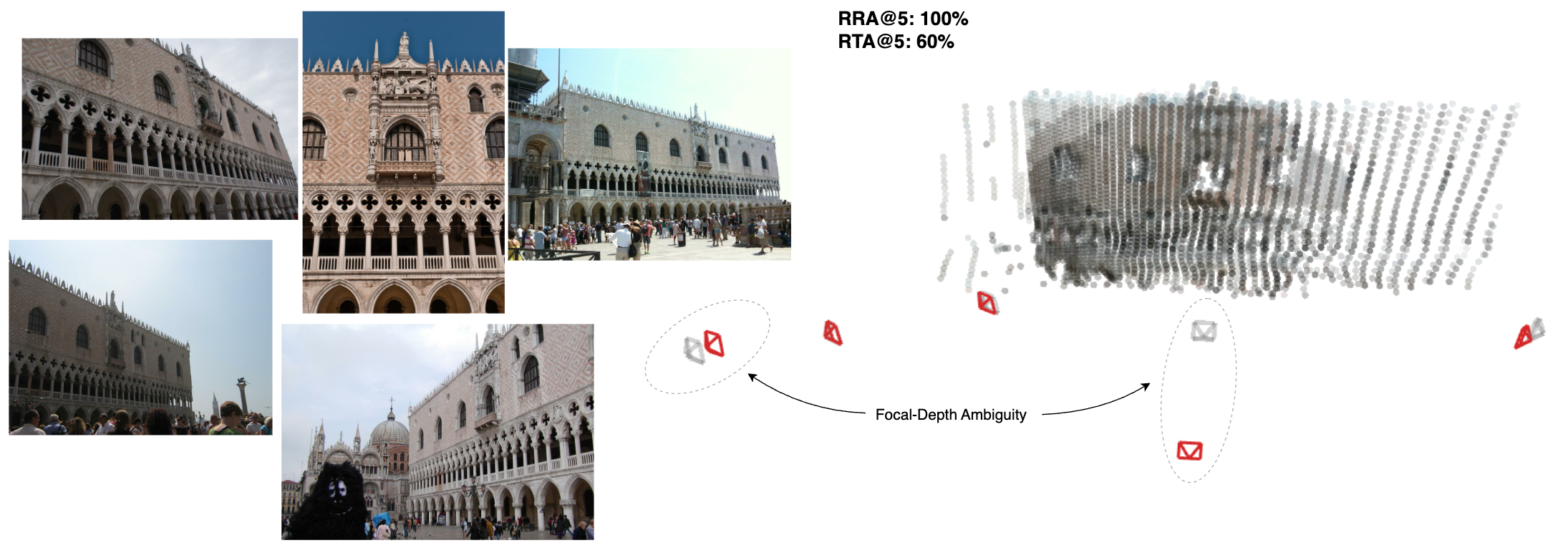}
    \caption{\textbf{Visualization of Focal-Depth Ambiguity.} Left: Input images. Right: Estimated poses (red), ground truth poses (grey), and the reconstructed scene.}
    \label{fig:focal_depth_am}
\end{figure}

\vspace{1mm}\noindent\textbf{Focal-Depth Ambiguity.}
As illustrated in Figure~\ref{fig:focal_depth_am}, our optimization occasionally suffers from scaling drift, yielding high Relative Rotation Accuracy (RRA@5$^\circ$) but comparatively low Relative Translation Accuracy (RTA@5$^\circ$). This stems from the focal-depth ambiguity in uncalibrated multi-view reconstruction. Under the pinhole model ($x = f \frac{X}{Z}$), scaling focal length $f$ and depth $Z$ by a factor $\alpha$ leaves the 2D projection identical ($\frac{\alpha f X}{\alpha Z} = f \frac{X}{Z}$). 

\begin{table*}[t]
\centering
\caption{\rev{\textbf{Additional baselines.} COLMAP/GLOMAP and their variants vs. ours.}}
\label{tab:more_baselines}
\begin{adjustbox}{max width=\linewidth}
\begin{tabular}{l | ccc | ccc | c | ccc | ccc | c | ccc | ccc | c | ccc | ccc | c}
\toprule
\multirow{3}{*}{Methods} & \multicolumn{7}{c|}{DTU\cite{Jensen_2014_CVPR}} & \multicolumn{7}{c|}{Mip-NeRF 360\cite{barron2022mipnerf360}} & \multicolumn{7}{c|}{ETH3D\cite{schops2017multi}} & \multicolumn{7}{c}{IMC2021\cite{jin2021image}} \\
\cmidrule(lr){2-8} \cmidrule(lr){9-15} \cmidrule(lr){16-22} \cmidrule(l){23-29}
 & \multicolumn{3}{c|}{RRA} & \multicolumn{3}{c|}{RTA} & \multicolumn{1}{c|}{\rev{AUC}} & \multicolumn{3}{c|}{RRA} & \multicolumn{3}{c|}{RTA} & \multicolumn{1}{c|}{\rev{AUC}} & \multicolumn{3}{c|}{RRA} & \multicolumn{3}{c|}{RTA} & \multicolumn{1}{c|}{\rev{AUC}} & \multicolumn{3}{c|}{RRA} & \multicolumn{3}{c|}{RTA} & \multicolumn{1}{c}{\rev{AUC}} \\
 & @1$^\circ$ & @5$^\circ$ & @10$^\circ$ & @1$^\circ$ & @5$^\circ$ & @10$^\circ$ & \rev{@10$^\circ$} & @1$^\circ$ & @5$^\circ$ & @10$^\circ$ & @1$^\circ$ & @5$^\circ$ & @10$^\circ$ & \rev{@10$^\circ$} & @1$^\circ$ & @5$^\circ$ & @10$^\circ$ & @1$^\circ$ & @5$^\circ$ & @10$^\circ$ & \rev{@10$^\circ$} & \rev{@3$^\circ$} & @5$^\circ$ & @10$^\circ$ & \rev{@3$^\circ$} & @5$^\circ$ & @10$^\circ$ & \rev{@10$^\circ$} \\
\midrule
COLMAP\cite{schonberger2016sfm} & \hcr{74.9} & \hcr{87.3} & \hcr{87.3} & \hcr{36.1} & \hcr{74.6} & \hcr{82.6} & \hcr{65.8} & \hcbr{100.0} & \hcbr{100.0} & \hcbr{100.0} & \hcr{98.2} & \hcbr{100.0} & \hcbr{100.0} & \hcur{97.9} & \hcbr{90.5} & \hcur{90.6} & \hcr{90.6} & \hcbr{80.1} & \hcur{89.2} & \hcur{90.1} & \hcur{84.8} & \hcr{81.9} & \hcr{85.3} & \hcr{88.4} & \hcr{54.9} & \hcr{65.3} & \hcr{75.4} & \hcr{58.9} \\
\rev{COLMAP + DKM\cite{edstedt2023dkm}} & \hcur{88.5} & \hcr{97.1} & \hcr{97.2} & \hcur{62.2} & \hcur{93.2} & \hcur{95.9} & \hcur{83.7} & \hcr{79.1} & \hcr{79.1} & \hcr{79.1} & \hcr{78.1} & \hcr{79.1} & \hcr{79.1} & \hcr{77.6} & \hcr{46.6} & \hcr{77.1} & \hcr{80.3} & \hcr{23.2} & \hcr{59.3} & \hcr{69.1} & \hcr{51.0} & \hcur{91.1} & \hcbr{94.3} & \hcur{96.5} & \hcr{63.3} & \hcr{75.3} & \hcbr{86.6} & \hcr{67.4} \\
\rev{COLMAP\cite{schonberger2016sfm} + SP/SG\cite{detone2018superpoint,sarlin2020superglue}} & \hcr{60.9} & \hcr{86.6} & \hcr{87.8} & \hcr{20.0} & \hcr{68.5} & \hcr{81.2} & \hcr{58.2} & \hcbr{100.0} & \hcbr{100.0} & \hcbr{100.0} & \hcr{96.7} & \hcur{99.9} & \hcbr{100.0} & \hcr{97.0} & \hcr{81.6} & \hcr{87.1} & \hcr{87.2} & \hcr{67.9} & \hcr{81.0} & \hcr{82.8} & \hcr{76.0} & \hcr{90.3} & \hcur{92.9} & \hcr{94.7} & \hcur{64.4} & \hcur{75.6} & \hcr{85.3} & \hcr{67.8} \\
\rev{GLOMAP\cite{pan2024glomap}} & \hcr{75.7} & \hcr{86.8} & \hcr{87.2} & \hcr{48.7} & \hcr{73.4} & \hcr{78.6} & \hcr{66.4} & \hcbr{100.0} & \hcbr{100.0} & \hcbr{100.0} & \hcur{98.5} & \hcbr{100.0} & \hcbr{100.0} & \hcbr{98.0} & \hcr{86.0} & \hcr{88.8} & \hcur{91.4} & \hcr{70.7} & \hcr{83.0} & \hcr{85.5} & \hcr{78.2} & \hcr{78.5} & \hcr{82.2} & \hcr{87.3} & \hcr{53.6} & \hcr{63.7} & \hcr{73.8} & \hcr{57.4} \\
\rev{GLOMAP + DKM\cite{edstedt2023dkm}} & \hcbr{92.1} & \hcbr{98.9} & \hcur{99.2} & \hcbr{78.3} & \hcbr{97.7} & \hcbr{98.1} & \hcbr{89.6} & \hcbr{100.0} & \hcbr{100.0} & \hcbr{100.0} & \hcbr{98.7} & \hcbr{100.0} & \hcbr{100.0} & \hcur{97.9} & \hcr{58.5} & \hcr{66.4} & \hcr{67.8} & \hcr{51.8} & \hcr{62.5} & \hcr{65.0} & \hcr{58.3} & \hcr{90.2} & \hcr{92.5} & \hcr{94.8} & \hcbr{66.5} & \hcbr{77.0} & \hcr{85.7} & \hcbr{69.2} \\
\rev{GLOMAP\cite{pan2024glomap} + SP/SG\cite{detone2018superpoint,sarlin2020superglue}} & \hcr{3.6} & \hcr{24.0} & \hcr{43.1} & \hcr{0.5} & \hcr{5.4} & \hcr{10.4} & \hcr{4.5} & \hcur{99.1} & \hcur{99.7} & \hcur{99.9} & \hcr{86.7} & \hcr{98.0} & \hcr{99.1} & \hcr{93.6} & \hcr{61.9} & \hcr{73.9} & \hcr{81.2} & \hcr{51.4} & \hcr{65.4} & \hcr{70.2} & \hcr{60.3} & \hcbr{91.6} & \hcbr{94.3} & \hcbr{96.8} & \hcur{64.4} & \hcr{75.5} & \hcur{86.2} & \hcur{68.0} \\
\textbf{\ours{}} & \hcr{88.1} & \hcur{97.9} & \hcbr{99.8} & \hcr{52.2} & \hcr{83.1} & \hcr{91.7} & \hcr{75.0} & \hcr{96.7} & \hcr{99.6} & \hcbr{100.0} & \hcr{92.2} & \hcr{98.5} & \hcur{99.2} & \hcr{94.0} & \hcur{86.3} & \hcbr{96.9} & \hcbr{99.0} & \hcur{77.4} & \hcbr{90.1} & \hcbr{92.4} & \hcbr{85.0} & \hcr{59.5} & \hcr{71.2} & \hcr{84.0} & \hcr{29.9} & \hcr{39.5} & \hcr{52.4} & \hcr{37.1} \\
\bottomrule
\end{tabular}
\end{adjustbox}
\end{table*}

This ambiguity is most severe in planar scenes ($Z \approx Z_0$), where insufficient depth variation deprives the optimizer of parallax needed to lock in true scale. This explains our IMC2021 results (Table~\ref{tab:aggregated_results_all_horizontal}), which is dominated by flat building facades. In such scenes, the planar projection still constrains viewing angles, \rev{so our rotations trail the baselines by far less than our translations do}. However, lacking 3D structure, translation remains ambiguous. The optimizer freely pushes cameras away along the optical axis while increasing focal length, minimizing reprojection loss, but distorting estimated distances and degrading RTA.

Incremental SfM methods like COLMAP mitigate this by initializing from an image pair with strong baseline parallax, rigidly locking in focal length and geometry via PnP. Because \ours{} performs global, gradient-based optimization from a cold start without this parallax-gated initialization, it is inherently more susceptible to this ambiguity on flat scenes.

\subsection{\rev{Additional Baselines}}
\label{sec:more_baselines}

\rev[a6res2]{SuperPoint/SuperGlue matches behave very differently in COLMAP and GLOMAP (\Cref{tab:more_baselines}): COLMAP still reaches RTA@5$^\circ$ 68.5 on DTU, whereas GLOMAP's global positioning collapses to 5.4, indicating that sparse detector-based matches over object-centric captures leave the view graph too weakly constrained for rotation averaging and global positioning to succeed.}
 
\end{document}